\documentclass[conference]{IEEEtran}
\usepackage[sort,nocompress]{cite}
\usepackage[bookmarks=true,breaklinks=true,letterpaper=true,colorlinks,citecolor=blue,linkcolor=blue,urlcolor=blue]{hyperref}

\usepackage{wrapfig}
\usepackage{pgfplots}
\usepackage{layouts, dblfloatfix}
\usepackage{amsmath,amssymb,amsfonts,pifont}
\usepackage{mathtools}
\usepackage{graphicx}
\usepackage{textcomp}
\usepackage{xcolor}
\usepackage{braket}
\usepackage{physics}
\usepackage{comment}
\usepackage{boxedminipage}
\usepackage{soul,color}
\usepackage{todonotes}
\usepackage{algorithm} 
\usepackage{algpseudocode}
\usepackage{float}
\usepackage{algpseudocode}
\usepackage{amsmath,amssymb,amsfonts}
\usepackage [autostyle, english = american]{csquotes}
\MakeOuterQuote{"}
\usepackage[utf8]{inputenc}
\usepackage{bm}
\usepackage[mathscr]{euscript}

\let\oldtodo\todo
\renewcommand{\todo}[1]{\oldtodo[inline, color=yellow]{{\bf TODO:} #1}}

\newcommand{\donotshow}[1]{}

\def\eg{{\em e.g.}, }
\def\ie{{\em i.e.}, }

\author{
    \IEEEauthorblockN{Anshujit Sharma, Matthew Burns, Andrew Hahn, and Michael Huang}
    \IEEEauthorblockA{Department of Electrical and Computer Engineering, 
    University of Rochester, NY 14627, USA \\
    anshujitsharma@rochester.edu, mburns13@ece.rochester.edu, ahahn7@ece.rochester.edu, michael.huang@rochester.edu}
}

\graphicspath{{Figs/}}
\title{Augmented Electronic Ising Machine as an Effective SAT Solver}

\begin{document}

\maketitle

\sloppy
\pagestyle{plain}

\begin{abstract}

With the slowdown of improvement in conventional von Neumann systems, increasing
attention is paid to novel paradigms such as Ising machines.
They have very different approach to NP-complete optimization problems. 
Ising machines have shown great potential in solving binary optimization problems like MaxCut. In this paper, we present an analysis of these systems in satisfiability (SAT) problems. We demonstrate that, in the case of 3-SAT, a basic architecture fails to produce meaningful acceleration, thanks in no small part to the relentless progress made
in conventional SAT solvers. Nevertheless, careful analysis attributes part of the failure to the lack of two important components: cubic interactions and efficient randomization heuristics. To overcome these limitations, we add proper architectural support for cubic interaction on a state-of-the-art Ising machine. More importantly, we propose a novel semantic-aware annealing schedule that makes the search-space navigation much more efficient than existing annealing heuristics. With experimental analyses, we show that such an \emph{Augmented} Ising Machine for SAT (AIMS), outperforms state-of-the-art software-based, GPU-based and conventional hardware SAT solvers by orders of magnitude. We also demonstrate AIMS to be relatively robust against device variation and noise.

\end{abstract}

\section{Introduction}
\label{sec:introduction}
 
As complexity of semiconductor fabrication technology reaches unprecedented levels,
there is an increasing urgency to seek other sources of improvement to speed and cost
of computing systems. Physics-based computing systems are getting increasing
attention from the architecture community. Quantum computing has some interesting
features such as changing how problem complexity scales with 
size~\cite{Shor_1997, grover1996fast}. 
Nevertheless, the technical challenges of building large-scale error-corrected
systems remain formidable.

Another area that has seen rapid recent progress is the subfield loosely termed
Ising machines. These machines are dynamical systems that can achieve extremely 
fast and energy-efficient search of an "energy landscape" defined by a parameterized
Ising model. This has the effect of optimizing the Ising model or
the equivalent of a QUBO (quadratic unconstrained binary optimization) form.
While the search can involve quantum mechanics (as done in quantum annealers
from D-Wave), it can also be done classically.

While functioning prototypes are only beginning to emerge, 
many Ising machine design concepts exist. For those that operate in the classical regime,
scaling in size appears less challenging than gate-based quantum computers.
For example, NTT recently developed a 100,000-node Coherent Ising Machine 
(CIM)~\cite{100k_cim}. Consequently, it is conceivable that these machines can
already outperform state-of-the-art sequential solvers. However, as we will
discuss in more detail, there are more subtle issues beyond just scaling mentioned 
above.

We first take an application-centric perspective to the analysis
of these machines: to what extent can they accelerate solving real-world
problems. One such well known problem that appears in many applications is
\emph{boolean satisfiability (SAT)}, which attempts to find a satisfying assignment
for a given logical statement~\cite{satisfiability_handbook}.
While Sec.~\ref{sec:background} discusses SAT in more detail, we note here that
it is an excellent benchmark: 
While many applications can be mapped into a QUBO problem, there are
nuances (\eg efficient navigation and cubic interactions); 
in the case of SAT problems, the nuances have important
implication on a machine's capabilities as we will discuss later.

The main contributions of this paper can be summarized
as follows. \ding{172} We demonstrate that standard Ising machines
are not competitive in solving SAT compared to
state-of-the-art SAT solvers. Our analysis show that the 
main causes are lack of support for high order
interaction and efficient search-space navigation.
\ding{173} Motivated by these limitations, we extend a baseline 
Ising machine to support high order terms. \ding{174} But more importantly, we
propose a new semantic-aware annealing heuristic to significantly
improve Ising machines'
navigational efficiency. \ding{175} We demonstrate that our
proposed "augmented" Ising machine is indeed orders of
magnitude faster than existing state-of-the-art SAT solvers.

For the convenience of the reader, a table of frequent acronyms follows:

\begin{table}[htp]\centering\scriptsize
\scalebox{0.9}{
\hspace{-9pt}
\setlength{\tabcolsep}{2pt}
\begin{tabular}{|l|p{1.88in}|} 
\hline
BRIM & Bistable Resistively-coupled Ising Machine   \\\hline
CDCL & Conflict-Driven Clause Learning              \\\hline
KZFD & Kolmogorov, Zabih, Freedman, Drineas         \\\hline
QUBO & Quadratic Unconstrained Binary Optimization  \\\hline
\end{tabular}   
\begin{tabular}{|l|p{1.04in}|}  \hline
CNF  & Conjunctive Normal Form  \\\hline
QC   & Quantum Computing        \\\hline
SA   & Simulated Annealing      \\\hline
TTS  & Time-To-Solution         \\\hline
\end{tabular}
}
\end{table}
\vspace{-9pt}
\section{Background and Related Work}
\label{sec:background}


\subsection{Boolean Satisfiability (SAT)}
\label{ssec:sat_section}
Boolean satisfiability has a large domain of 
application including circuit hardware verification~\cite{circuit_cbmc}, 
cryptographic protocols~\cite{sat_security}, software verification~\cite{sat_software}, 
hierarchical planning~\cite{sat_planning}, mapping~\cite{sat_multiprocessor}, 
scheduling~\cite{sat_scheduling}, and even
hardware security attacks \cite{bit_sat_attack, cyc_sat_attack}. 

A generic SAT problem asks whether a propositional logic
formula \emph{F} is satisfiable for some mapping of variables to truth
values~\cite{satisfiability_handbook}. \emph{F} consists of boolean-valued
variables acted on by unary logical operator (\emph{negation}: $\lnot$) or
binary ones (\emph{and}: $\land$; \emph{or}: $\lor$; and \emph{implications}:
$\to$, $\iff$) (\eg $F=x_1\lor (\lnot x_2)\land (x_3\to x_4)$)\footnote{For compactness, $\bar{x}$ is
generally used instead of $\lnot x$.} The statement is typically expressed in
Conjunctive Normal Form (CNF)~\cite{satisfiability_handbook}, \ie
a logical conjunction of separate \emph{clauses} (\eg $F=C_1\land C_2 \land
C_3$). Each clause is a logical disjunction of a number of \emph{literals}.
Literals are boolean variables or their negations (\eg $C=x_1 \lor \overline{x_2}
\lor x_3$). Propositional logic statements can be converted to CNF in
linear time by introducing new variables and clauses ~\cite{Tseitin1983}.
 When the number of literals in any clause is no more than $k$, the problem 
is called $k$-SAT. When $k\ge 3$, the problem is NP-complete.
3-SAT can be used to formulate any NP-complete problem in
standard CNF format, with SAT solvers as an efficient means of
solution~\cite{sat_np_complete, cook_levin_paper}.

\subsection{Software SAT solvers}
\label{sec:sw_sat}
In the worst case, the deterministic evaluation of a SAT problem requires
enumeration of all possible variable truth values. Correspondingly, a naive
deterministic SAT algorithm is $O(2^N)$ for an $N$-variable formula. Hence,
modern SAT solvers rely on partial search-space exploration to attempt to find
satisfying solutions. 
While a more comprehensive review can be found in 
\cite{satisfiability_handbook}, the more relevant point to this paper is that
there are two primary paradigms for SAT algorithms.

Most contemporary SAT solvers are based upon the Conflict-Driven Clause
Learning (CDCL) algorithm \cite{grasp_cdcl} (an extension of DPLL algorithm
\cite{dpll}). CDCL is a sequential approach in which variables are
incrementally assigned values until either all variables are assigned (SAT) or
a conflict is reached (all literals of a clause are false). Conflicts cause
search-space backtracking. Modern incarnations rely heavily on heuristics
to trim the search space and prioritize promising decision branches. KISSAT
\cite{kissat_github} is a prime example. Its variants have won the SAT solver
competition from 2020 to 2022 ~\cite{sat_competition,kissat_21}.

The second class are known broadly as "stochastic local search" (SLS)
algorithms, which perform localized random walks over the search
space. A paradigmatic example is WalkSAT~\cite{walksat}, an extended version
of Sch\"{o}ning's algorithm \cite{schoning}.
Rather than explicitly trimming the search
space, SLS algorithms attempt to maximize the probability of encountering a
satisfying solution with stochastic mechanisms. WalkSAT has 
an expected time complexity of $O(1.334^N)$ -- a polynomial improvement over
$O(2^N)$~\cite{schoning}, which has later been improved to $O(1.308^N)$ with the PPSZ algorithm ~\cite{ppsz}. 
SLS algorithms run until either a SAT is found or a timeout.
Therefore, they cannot refute a CNF formula and are classified as \emph{incomplete}
solvers, without a \emph{guarantee} of success.
Note that in practice, even complete solvers 
are generally used with a timeout to prevent excessive runtime from full exponential search space exploration.


Hybrid solvers mixing CDCL and SLS have been
 shown to be highly effective~\cite{sat_competition, sat22}. The close analogy between Ising machines and SAT heuristics was also exploited in a hybrid  CDCL/Ising machine-based approach \cite{tan2023hyqsat}.

\subsection{Conventional hardware SAT solvers}


Hardware SAT solvers can also be classified by its general algorithm into CDCL or SLS styles.
The former is further separated into coprocessors 
(\eg handling only \emph{boolean constraint propagation}~\cite{davis_bcp})
and dedicated solvers (\eg \cite{gulati_cdcl_solver}). 
While a 2017 survey~\cite{sat_hw_survey} contains a good summary of earlier works, 
a few recent works have focused on SLS style algorithms such as
probSAT~\cite{probSAT} and AmoebaSAT~\cite{amoebasat}. 
These algorithms are easier to implement in hardware and expose more parallelism
leading to orders of magnitudes of speedups over the respective software version~\cite{brw_mtj, amoebasat_fpga, amoebasat_fpga_original}. Note that as discussed earlier, software speeds vary
dramatically. As a result, speedups over a particular software solver is
not a conclusive evidence.
Finally, like their software counterpart, SLS-style hardware
lack the completeness guarantees. 

\subsection{Optimization using quantum computing}
\label{ssec:qc}

Quantum computing (QC) methods have been previously proposed for NP-Complete problems. There are two broad, theoretically equivalent categories of QC~\cite{qaa_equiv_uqc}: \ding{172} gate-based QC or "Universal Quantum Computing (UQC)" and \ding{173}  "Adiabatic Quantum Computing (AQC)". UQC algorithms perform sequences of unitary operations on quantum bits (qubits) before sampling their state. Grover's algorithm for unstructured search \cite{grover1996fast} is a popular example. However, error sensitivity in the "Noisy-Itermediate Scale Quantum" (NISQ) era limit its immediate usefulness \cite{grover_nisq}. Noise-resilient, "shallow" algorithms have been proposed for SAT as well \cite{nisq_2022}. Examples are generally hybrid algorithms, either classically optimizing a parametric quantum circuit \cite{qaoa_original, qaoa_sat} or using limited QC subroutines \cite{3qsat}. However, these proposals have either used noiseless simulations \cite{qaoa_sat} or were purely theoretical \cite{3qsat}, so their real-world applicability is unknown.

AQC algorithms evolve qubits under a mixture of two non-commuting Hamiltonians: a "simple" operator with a known ground state and a problem-specific cost function. Ideally, one begins in the ground state of the simple Hamiltonian and slowly transitions to the other. The adiabatic theorem provides ground state guarantees for sufficiently slow transitions \cite{aqc}. Adiabatic algorithms for SAT have been proposed which theoretically promise high-quality solutions \cite{adiabatic_sat_orig, adiabatic_sat}. However, NISQ-era noise precludes adiabaticity \cite{nisq_2022}. A relaxation is "quantum annealing" (QA): retaining mixed operators but abandoning theoretical guarantees \cite{qa_original}. In QA, the transverse field decreases over the course of system evolution similar to classical simulated annealing \cite{simulated_annealing}. QA was shown to satisfy 24/100 3-SAT instances.\footnote{This improved to 99/100 with post-processing~\cite{qa_sat_quality}, although the post-processing took longer than a brute-force enumeration.} Testing was limited to 10-variable problems due to QA topology constraints \cite{qa_sat_quality, qa_sat_routing} as we will discuss in
Sec.~\ref{ssec:embedding}.

Throughout the analyses, we will report the performance of an optimistic estimate
of a near-term quantum algorithm. As we will see, such approaches do not seem to 
offer any benefit in the NISQ era.

\subsection{Ising machines}
\label{ssec:ising_machines}

The Ising model was originally used to describe 
the Hamiltonian ($H$) for a state of spins on some lattice.
A modified version (Eq.~\ref{eq:ising})
of the formula is more useful in the discussion of optimization
problems. 
\begin{equation}
    \begin{split}    
        H(\sigma) &= -\sum_{(i,j)}J_{ij}\sigma_i\sigma_j - \sum_{i}h_i\sigma_i 
            = -\sigma^{\top}J\sigma-h^\top\sigma
        \label{eq:ising}
    \end{split}
\end{equation}
A loose physical analogy is as follows. A system of spins
($\sigma_i \in \{-1, +1\}$) are subject to the influence of 
a set of parameters $J_{ij}$ and $h_i$. $J_{ij}$ describes the coupling between
two spins ($\sigma_i$ and $\sigma_j$), while $h_i$ describes the
influence of an external magnetic field on spin $\sigma_i$. Given this setup, the
spins will naturally gravitate towards the lowest energy state (ground state).
Technically, the probability of each state follows the Boltzmann distribution:
\begin{equation}
    P(\sigma) \propto e^{-\frac{H(\sigma)}{T}}
    \label{eq:boltzmann_dist}
\end{equation}
If a physical system of spins can be set up subject to the programmable
parameters, then we gradually lower the temperature ($T$) of the system (annealing); when it
approaches 0, the probability of the system in ground state(s) will approach 1. 
Such a physical system is thus effectively performing an optimization (where
the objective function is Eq.~\ref{eq:ising}). In practice, different types
of physics are involved to achieve the effective optimization. These physical
systems are often referred to as Ising machines and have been shown to find good
solutions to combinatorial optimization problems (COP) at extraordinary 
speeds~\cite{100k_cim, inagaki2016cim, cim_yamamoto, McMahon614, 16bit_cim, 
ncomm_bohm2019, wang2019oim, afoakwa.hpca21}.

In principle, all NP-complete problems can be converted into a COP
and thus benefit from the speed and efficiency of an Ising machine. 
Indeed, \cite{lucas2014ising} is often cited for this purpose. 
In practice, however, there are subtle issues in
the process of solving a problem with an Ising machine: \begin{enumerate}
    \item A "native" formulation of the problem may contain cubic or higher-order 
    terms. Converting them down to quadratic form increases the number of spins needed.
    \item In many prototypes, coupling is only limited to between nearby spins. This
    means even more auxiliary spins are needed.
    \item The navigation through the energy landscape is purely based on the formulation.
    Any problem-specific insight may be lost.
\end{enumerate}

In response to the first issue, a number of proposals for "high order" Ising models have recently been proposed. High order models expand the Ising Hamiltonian to arbitrary polynomial degree. Efforts so far have focused on 3\textsuperscript{rd} and 4\textsuperscript{th} order interactions, with applications both to 3-SAT and its variants ~\cite{bashar2023oscillator, high_order_ising_oim}, and hypergraph partitioning \cite{oim_hypergraph}. Ising models generalized to $K$-degree interactions are of the form \cite{chermoshentsev2021polynomial}:
\begin{equation}
    \begin{split}    
        H(\sigma) = &-\sum_{i}h_i\sigma_i-\sum_{(i,j)}J^{(2)}_{ij}\sigma_i\sigma_j - \sum_{(i,j,k)}J^{(3)}_{ijk}\sigma_i\sigma_j\sigma_k \\
        &\dots -\sum_{(i,j,...,m)}J^{(K)}_{ij...m}\sigma_i\sigma_j...\sigma_m
        \label{eq:highising}
    \end{split}
\end{equation}

High order Ising Hamiltonians can be applied to a broader class of problems, collectively denoted as "polynomial unconstrained binary optimization" (PUBO) \cite{chermoshentsev2021polynomial, tari2022pubo} or "higher-order unconstrained binary optimization" (HUBO) \cite{jun2023hubo, norimoto2023quantum}.

Simulation of high order Ising machines have been shown to outperform the respective baseline Ising machine solvers. The GPU/FPGA based simulated bifurcation algorithm, originally limited to quadratic systems, has been modified to allow for more general Hamiltonians \cite{kanao2022simulated}. An optical
Ising machine-inspired algorithm demonstrated solution quality superior to existing algorithms, though without significant speedup \cite{chermoshentsev2021polynomial}. 

Amongst nature-based Ising machines, oscillator-based dynamical models have been among the first demonstrated. 3\textsuperscript{rd} order models were proposed and modeled for classic 3-SAT and "Not-All-Equal" (NAE) 3-SAT problems, with successful demonstration on (tiny) 6-variable problems \cite{bashar2023oscillator}. An LC oscillator-based model was shown in simulations to solve 3-SAT problems of up to 250 variables, though with poor success rate of 2\%\cite{high_order_ising_oim}. 


So far, high order Ising machines proposed only added high order couplings. As we will discuss in Sec.~\ref{sec:motivation}, adding support for cubic interaction alone is insufficient for Ising machines to become competitive as a SAT solver. As discussed in Sec. \ref{sec:sw_sat}, there exists a huge amount of background work on SLS for SAT (PPSZ being the fastest known). These solve SAT much more efficiently (visiting less phase points, as we shall see) than an Ising machine that relies on a problem-agnostic
annealing schedule. In the next section, we will discuss the challenges in more detail which will motivate us to propose a fast, high order Ising machine that uses a novel annealing schedule. 
\section{Binary Optimization Approach to SAT}
\label{sec:motivation}

In an $N$-variable unconstrained binary optimization (UBO) problem, there are
$2^N$ possible states $(x_1, x_2, ... , x_N), x_n \in \{0, 1\}$. Each state is
associated with a value of an objective function. The value can be thought of
as a cost or energy to be minimized. If the cost function is quadratic or
higher-order polynomial, the problem is called QUBO or PUBO, respectively. Given
that each state (or \emph{phase point}) has an energy level, the
entire phase space can be thought of as an energy \emph{landscape}. Consider a
3-SAT problem in CNF: $F = \bigwedge_{i=1}^{_M}(\ell_{i1} \lor \ell_{i2} \lor \ell_{i3})$,
where each term $\ell_{ij}$ is a literal $x_n$ or $\bar{x}_n$. We can simply
define the energy for each phase point as the number of unsatisfied clauses 
of that phase point. If we
use 1 and 0 to represent \emph{true} or \emph{false} of each Boolean variable
$x_n$, then the energy (or the objective function to be minimized) is
\begin{equation}\begin{split}
        H  =  \sum_{i=1}^M  (1-\ell_{i1})(1-\ell_{i2})(1-\ell_{i3});\\
        \underline{ e.g.,\ (x_7\lor \bar{x}_{11}\lor x_{12}) \xrightarrow{} (1-x_7)x_{11}(1-x_{12})}.
   \label{eq:pubo_3sat}
\end{split}
\end{equation}
Given this energy landscape, a generic algorithm (\eg simulated annealing 
or SA) can thus be used as an optimizer. An Ising machine can be thought of
as a potentially very efficient alternative to von Neumann hardware running SA. 
However, there are 3 issues mentioned in Sec.~\ref{ssec:ising_machines}.
We now discuss them with more quantitative detail.

\subsection{Cost of QUBO formulation}
\label{ssec:cubic_vs_qubo}

\begin{figure}[htb]
    \centering
    \includegraphics[width=0.33\textwidth]{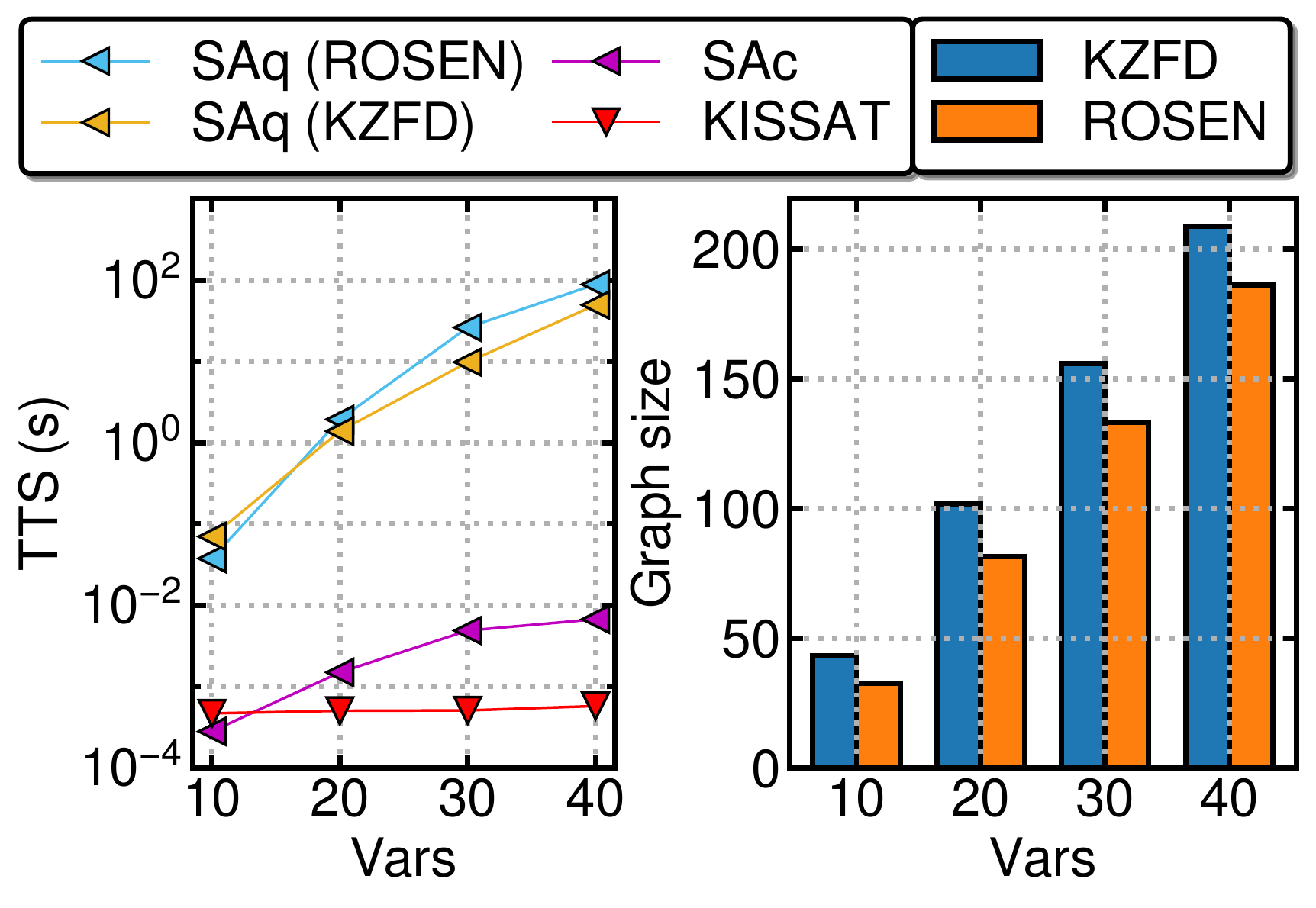}
    \caption{[Left] Time-to-solution for simulated annealing using cubic and quadratic 
    (with Rosenberg and KZFD rules) objective functions. For reference KISSAT is also
    included. [Right] The average size of QUBO graphs using KZFD and
    Rosenberg's formulations for different problem sizes.}
    \label{fig:pubo_vs_qubo}
\end{figure}

While we use binary objective functions because of their straightforward connection
with Boolean logic, we note that 
the standard Ising formulation (Eq.~\ref{eq:ising}) is completely convertible
to and from QUBO formulations, by simple conversion between spin
and bit representations ($\sigma_i = 2x_i-1; \sigma_i \in \{-1, 1\}$,
$x_i \in \{0, 1\}$).\footnote{There is a small technicality though: Due to its
physics origin, the Ising formulation has no self-coupling; In other words,
$\forall i, J_{ii}=0$ in Eq.~\ref{eq:ising}. In a typical QUBO formulation
($\sum W_{ij}x_ix_j$), there is no such constraint. Indeed, because in binary
$x_i^2=x_i$, $W_{ii}$ helps eliminate the need for coefficient of linear terms
for compactness of representation. In the Ising formulation, the linear 
coefficients $h_i$ correspond to $W_{ii}$ in QUBO.}  However, 
as can be seen in Eq.~\ref{eq:pubo_3sat},
the objective function has cubic terms $\ell_{i1}\ell_{i2}\ell_{i3}$ and is thus
not a QUBO formula. A standard
practice is to reduce such an objective function to quadratic form (QUBO) at
the expense of extra auxiliary variables. Perhaps the most widely used approach (due to 
Rosenberg ~\cite{rosen}) is to add a new variable $z$ as a substitute for $x_1x_2$
and introduce a penalty when $z\neq x_1x_2$. With this substitution, the
cubic term is replaced by a purely quadratic expression:
    \begin{equation}
        \begin{split}
            \pm x_1x_2x_3 &\rightarrow \pm zx_3 + k(x_1x_2 - 2(x_1+x_2)z + 3z),
        \end{split}
    \end{equation}
where $k>1$.
This seemingly small change has a significant impact. Every clause will generate
an extra spin, leading to a total of $M+N$ spins for an $N$-variable, $M$-clause
problem. Unfortunately, for the most difficult SAT problems,
$M$ is usually around $\alpha_c=4.17$ times that of $N$~\cite{nishimori.book2001}.
This means the number of spins will roughly quintuple and the size of the phase space will
grow from $2^N$ to $2^{5N}$. After carefully analyzing how the landscape reacts to the
penalty, we realized that the Rosenberg substitution can be improved.
As it turns out, our improved version is merely reinventing the wheel
of KZFD~\cite{KZFD, FD}. This formulation has two different
substitutions depending on the sign of the cubic term (Eq.~\ref{eq:kzfd}). 
    \begin{equation}
        \begin{split}
            -x_1x_2x_3 &\rightarrow (2-x_1-x_2-x_3)z\\
            +x_1x_2x_3 &\rightarrow (1+x_1-x_2-x_3)z + x_2x_3
        \end{split}
        \label{eq:kzfd}
    \end{equation}
To show the impact of quadratization, we will show how the solution
time grows (dramatically) as the problem size increases. To more
accurately reflect solution time (since the algorithm is stochastic),
we use Time-To-Solution (TTS)~\cite{TTS}, defined as
the expected time to find a solution with 99\% probability.
Over multiple instances with runtime of $T_{run}$ and a success 
probability of $P$, 
TTS can be calculated using Eq.~\ref{eq:tts}.
\begin{equation}
    TTS = \begin{cases}
        T_{run}\times\dfrac{\log_{10}(0.01)}{\log_{10}(1-P)} & P < 0.99\\
        T_{run} & \textrm{otherwise}
    \end{cases}
    \label{eq:tts}
\end{equation}
 We plot TTS of SA using both cubic formula (SAc) directly (Eq.~\ref{eq:pubo_3sat}) and the QUBO translation (SAq) with the KZFD and Rosenberg rules in Fig.~\ref{fig:pubo_vs_qubo} left. 

 Clearly, adding spins is not just costly in hardware, it has a significant impact on the annealer's ability to navigate the landscape to reach a certain solution quality. The outperformance of SAc over SAq shows we can avoid costs due to quadratization if
the hardware can provide cubic interaction. 

Fig.~\ref{fig:pubo_vs_qubo} right shows the size bloat in both methods of conversion.
We can see that to a first approximation, both SAq's are slower due to significantly
larger phase space. Secondarily, when cubic interaction is absent, the KZFD formulation scales slightly better despite having a slightly larger size penalty. 
This is due to a specific advantage of the resulting energy landscape 
the detail of which need not concern us in this paper. 

\noindent \underline{\bf Observation 1: Cubic interaction is very useful.}

\subsection{Topological limitations}   
\label{ssec:embedding}

On top of translation into a QUBO formula, an Ising machine may impose another
constraint to the problem formulation and in turn requires additional auxiliary
spins. This is the topology of the coupling array. A generic matrix $J$
(Eq.~\ref{eq:ising}) may require any two spins to be coupled. But practical
systems have commonly offered only near-neighbor coupling\cite{pelofske.icrc.2019,
oim_128_prototype}. One reason is that having fewer couplers may allow a (nominally)
bigger system under the same hardware budget. But is the trade-off worthwhile?
Let us look at a concrete example of Chimera $k_{4,4}$ used in D-Wave 2000Q.

If the problem has a dense coupling matrix, there is no advantage, only disadvantage. And  the disadvantage is both clear and substantial. For an $N$-node all-to-all problem, $O(N^2)$ couplers are needed. With a Chimera topology, we cannot avoid the $O(N^2)$ couplers (in fact we need more). In addition, having $O(N^2)$ couplers automatically means $O(N^2)$ nodes. With more than 2000 nominal spins, the 2000Q can only map a dense problem of size $\leq 64$, assuming all qubits are stable and available. Compared to an all-to-all topology, the Chimera topology ends up using twice as many couplers and 23 times more nodes.  Not only is the topology more costly, it fundamentally complicates landscape navigation. 

\begin{figure}[htb]
    \centering
    \includegraphics[width=0.41\textwidth]{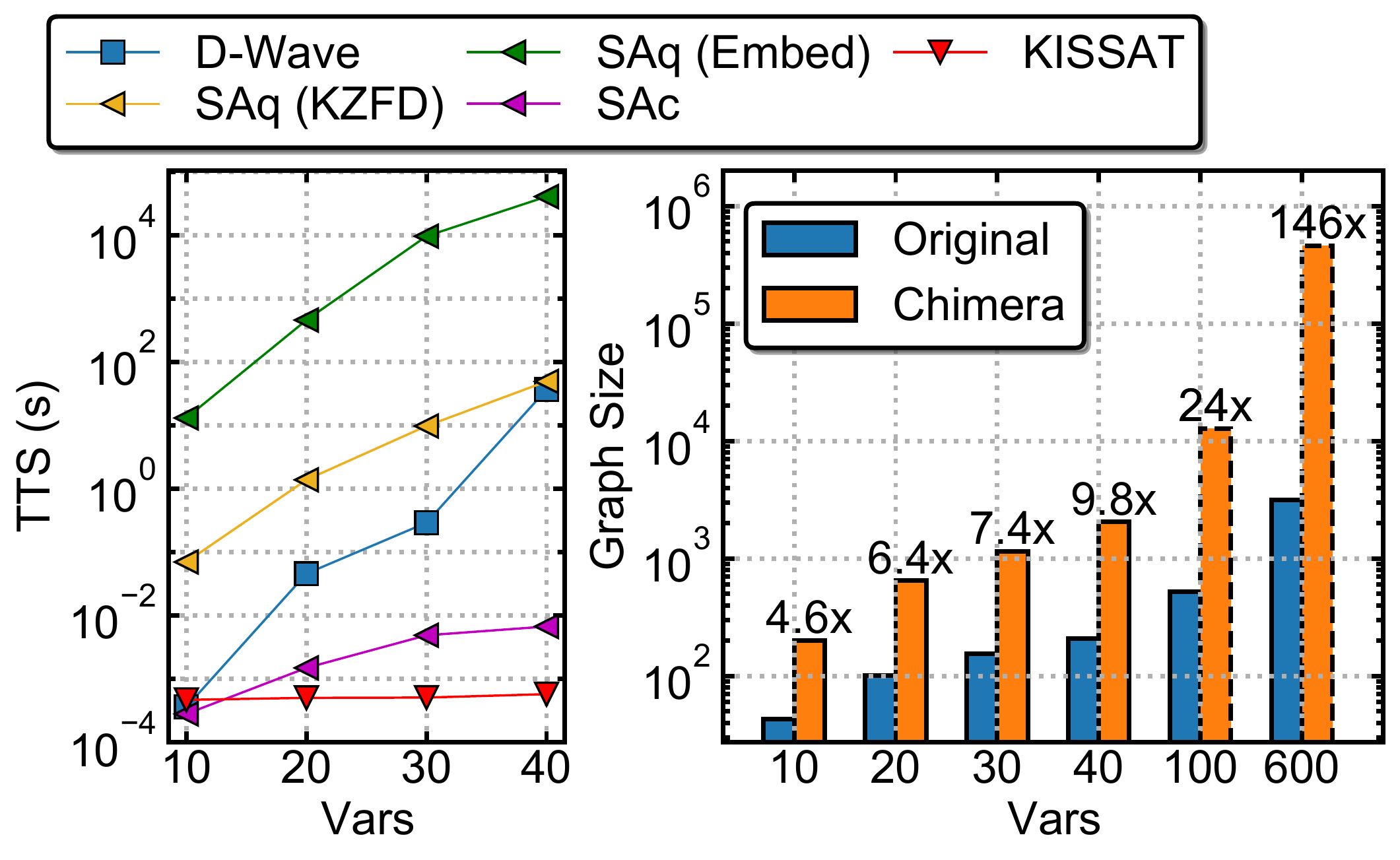}
    \caption{[Left] Performance of D-Wave, SA running original cubic formula (SAc), quadratic translations SAq (KZFD), and after 
    embedding to $k_{4,4}$. KISSAT is added for reference.
    [Right] Size of the minimum square $k_{4,4}$ Chimera lattice needed for embedding with the \texttt{minorminer} tool~\cite{minor.embedding}
    compared with the original KZFD graphs. The minimum Chimera sizes for 100 and 600 variables were extrapolated from the observed quadratic scaling. }
    \label{fig:annealing_comparison}
\end{figure}

For the 3-SAT problem, the situation is a bit better. The largest problem 2000Q can map (using KZFD formula) is about 42 variables and 178 clauses. At this size, the coupling matrix has a 4\% density. Compared to an all-to-all topology (which would have costed about 210 nodes and 22K couplers), Chimera allows about 3x (2K nodes, 6K couplers) hardware savings. However, the price paid for the savings is again clear and significant. \begin{enumerate} 
    \item  A minor embedding process is needed to convert the original problem into one that is compatible to the machine's topology. This process takes  non-trivial extra time. Embedding a 40-variable, 170-clause problem using the D-Wave \texttt{minorminer} heuristic takes over 17 seconds on average: several orders of magnitude more than the annealing time. 
    
    \item  With about 10x more spins, the transformed problem has a significantly larger phase space. A natural consequence is lower solution quality (recall the discussion in Sec.~\ref{ssec:cubic_vs_qubo}). To show that the phase space bloat is a major contributing factor, we use an SA solver and compare 3 different objective functions: cubic (Eq.~\ref{eq:pubo_3sat}), QUBO conversion using KZFD, and that with additional minor embedding for Chimera graph. As Fig.~\ref{fig:annealing_comparison} suggests, with every kind of phase space bloat, SA's performance is significantly degraded.  
\end{enumerate} 

Based on our analysis, the take-away point is: for the costs in
solution quality and extra processing time, 
it is not clear a near-neighbor topology such as chimera is providing much
benefit even for the extremely sparse problem such as 3-SAT.

\noindent \underline{\bf Observation 2: Near-neighbor coupling is costly.}

\subsection{Navigational efficiency}   
\label{ssec:navigation}

It is clear that if the Ising machine hardware can provide support to
avoid adding many extra spins and the concomitant phase space bloat,
the performance will improve. But can it outperform a conventional 
software algorithm? That depends on the problem and the algorithm.
On the one hand, for a generic problem simply defined as an energy 
landscape by, say, Eq.~\ref{eq:ising}),
the answer is clearly yes. This is because our best algorithm is 
some variant of SA (\eg \cite{isakov}). SA navigates the energy landscape
purely by the formula without any additional insight. Our empirical
observations suggest that to reach similar solution qualities, both hardware
Ising machines and SA need to visit similar number of phase points;
and the former can be orders of magnitude faster and more efficient than
the latter to visit the same number of phase points.

On the other hand, for problems such as SAT where the landscape is
\emph{derived}, insights from the original problem formulation may
not be obvious in the cost function. Software algorithms may
explore these insights and arrive at powerful, problem-specific
heuristics to more efficiently navigate the phase space. Thus, what
software lacks in speed when visiting/evaluating each phase point
can be compensated by its "navigational efficiency", or arriving
at the solution by effectively visiting far fewer phases points.
We have already seen this in Fig.~\ref{fig:annealing_comparison}: while
SAc was faster than KISSAT for smaller problems, its advantage disappeared
as the problem size grew. This slowdown is to some extent explained by
navigational efficiency. The assignment propagation in CDCL 
has the effect of ruling out entire subspaces without explicitly visiting 
each phase point. Depending on the problem type and size, this can
help CDCL style algorithms over SLS.

\noindent\underline{\bf Observation 3: Navigational efficiency is important.}

\subsection{Summary}

\begin{wrapfigure}{r}{0.30\textwidth}
    \vskip -8pt
    \includegraphics[width=0.30\textwidth]{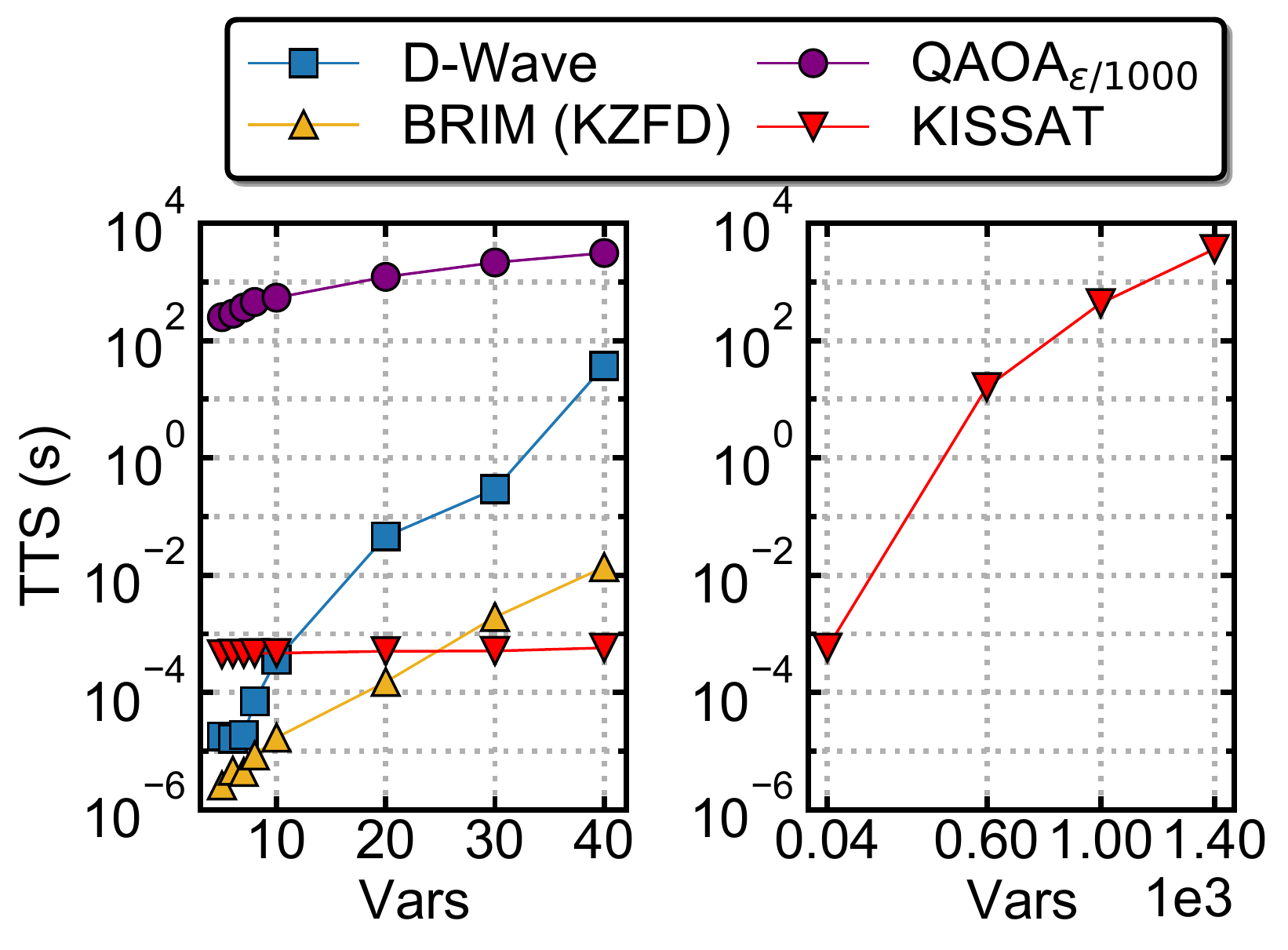}
    \vskip -8pt
    \caption{[Left] Execution times of different machines
    compared with a state-of-the-art SAT solver (KISSAT) for
    problems with different number of variables (Vars).
    [Right] Running KISSAT with problem sizes whose TTS matches
    that of QAOA and D-Wave at 40 variables.}
    \label{fig:brim_sat}
\end{wrapfigure}

To recap: special-purpose hardware such as Ising machines have strengths over conventional software running on von Neumann hardware. But they also face unique engineering challenges and lack the cumulative innovation invested in the state-of-the-art algorithms. Take the SAT problem as a case study, we can see the effect in Fig.~\ref{fig:brim_sat}.

In this figure, we model the speed of BRIM~\cite{afoakwa.hpca21}, D-Wave 2000Q, and KISSAT, each representing the state of the art in electronic Ising machine, QA, and software solvers. We also plot the optimistic simulation results of Quantum Approximate Optimization Algorithm (QAOA$_{\epsilon/1000}$) with 1000x lower noise levels ($\epsilon$) than today's quantum computers. We observe that quantum computing has massive overhead which makes it unlikely to be useful for SAT in the NISQ era. Now looking at BRIM's results, we see that just like D-Wave 2000Q, BRIM's early speed advantage quickly disappears as the problem size increases. In their present form, BRIM and D-Wave annealers are unable to be helpful for SAT. Moreover, based on TTS for current software, the minimum for special purpose hardware to become useful is probably around 600 variables.  For QA with near-neighbor topology, this translates to around 460,000 qubits. For QAOA, the threshold is "almost noiseless". In either case, these are goals beyond the NISQ era.

\donotshow{
\begin{figure}
    \centering
    \includegraphics[width=0.33\textwidth]{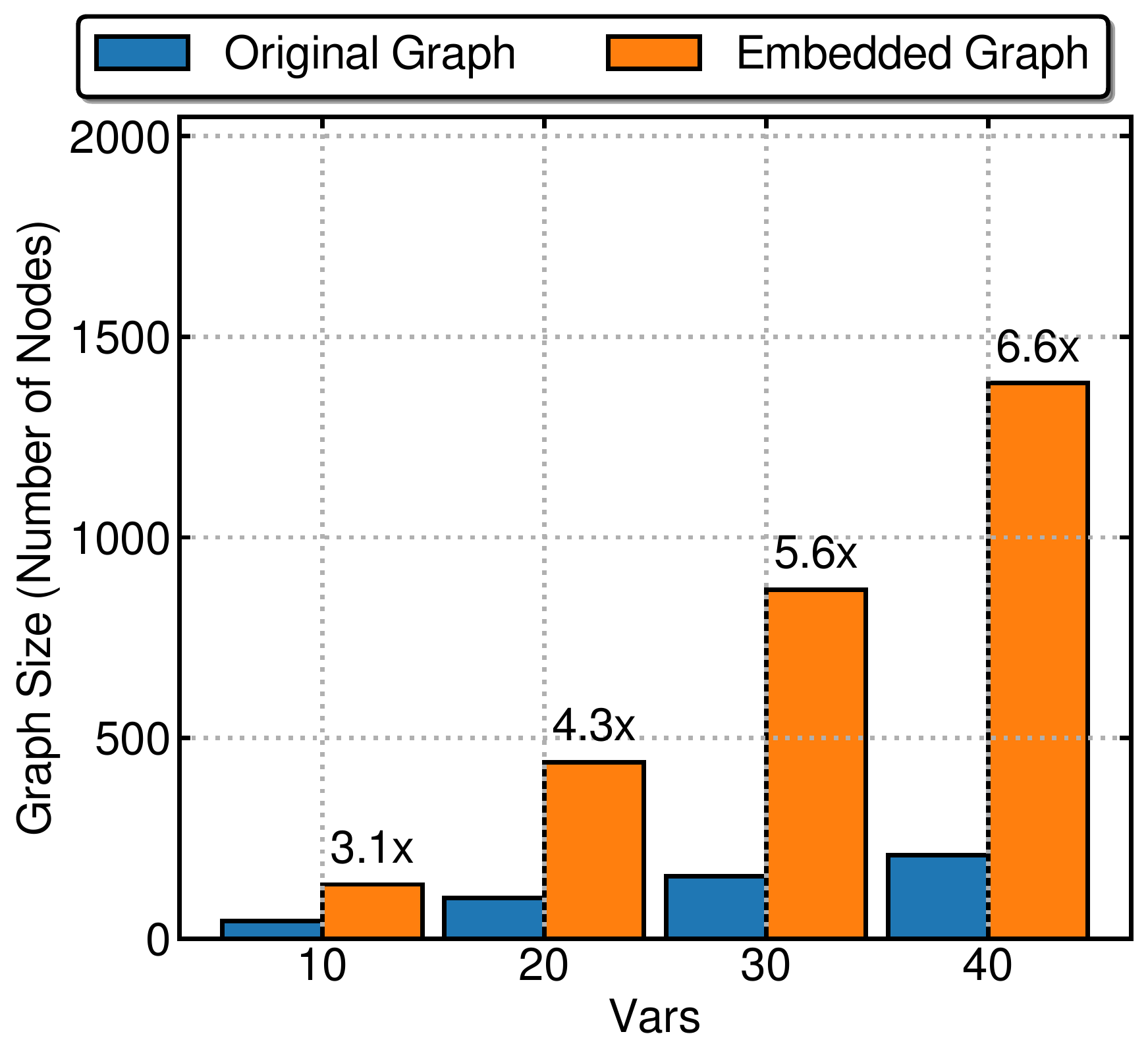}
    \caption{The original and embedded 2000Q Chimera graph sizes compared. The size of the embedded graph relative to the original is shown above the embedded column. The embeddings were done with the $\texttt{minorminer}$ $\texttt{find\_embedding}$ heuristic \cite{minor.embedding}.}
    \label{fig:embed}
\end{figure}
}



\section{Architectural Support}
\label{sec:arch}


Motivated by the observations so far, we propose a number of architectural supports that allow an electronic Ising machine to regain problem solving efficacy for SAT.

\subsection{BRIM as a baseline}
The Bistable Resistively-coupled Ising Machine (BRIM) \cite{afoakwa.hpca21} has three properties 
that make it an ideal starting point for our architectural exploration:
\begin{enumerate}
    \item \emph{CMOS-compatible}: This allows for an 
    easy integration with additional architectural support or ultimately
    with other types of processing elements in an SoC. Moreover,
    the voltage-based spin representation is easy to work with
    for additional digital logic, in contrast to more complicated
    spin representations such as qubits~\cite{PhysRevB.82.024511}, 
    or oscillator phases~\cite{inagaki2016cim}.
    
    \item \emph{Efficient dynamical system:} Unlike digital annealers that essentially 
    hard-wire a von-Neumann algorithm~\cite{STATICA, tatsumura.fpl19},
    BRIM is a dynamical system with programmable resistive coupling for 
    nano-scale capacitive spins. It is thus much more efficient due
    to fundamental reasons. In the original paper \cite{afoakwa.hpca21},
    the authors have demonstrated significant speedup for BRIM compared to a
    myriad of existing Ising machines.

    \item \emph{All-to-all topology}: Motivated by observation 2, we
    opt for an all-to-all topology (for quadratic interactions). With BRIM,
    this is both easy to achieve (all nodes are simply interconnected via
    a matrix of coupling units) and quite inexpensive in terms of
    hardware resources (each coupling unit accounts for only 1 $\mu m^2$ chip
    area with a 45nm Cadence PDK). Moreover, in a recent paper \cite{myisca22},
    the authors have demonstrated a multiprocessor BRIM that
    can solve a problem with over 16,000 nodes (all-to-all 
    connected) orders of magnitude faster than any existing solution.
\end{enumerate}

\subsection{A bit-based implementation of BRIM}
\label{ssec:baseline_arch}
We modify the baseline BRIM design \cite{afoakwa.hpca21} to represent
 $\{0,1\}$-bits instead of $\{+1,-1\}$ spins.
This is because for SAT, the bit representation is more direct and obviates
the need for QUBO-to-Ising conversion.

The blue module in Fig. \ref{fig:highlevel} shows the new baseline BRIM design
with bit representation.
Here, each node represents a variable $x_n$ in the 3-SAT formula, and each
coupling unit $q_{nj}$ and $l_n$ represents the quadratic and linear 
interactions respectively. The coupling resistance is given by
$R_{nj}=\frac{R}{|q_{nj}|}$, where $R$ is a constant and $q_{nj}$ is the 
coupling between nodes $n$ and $j$. The sign of the coupling is interpreted as
follows: \ding{172} if $q_{nj} > 0$, then the nodes are connected in parallel
(output of a node is connected to the positive input terminal of the other node 
via resistance $R_{nj}$ : $Out_n\rightarrow IN^+_j$, $Out_j\rightarrow IN^+_n$).
\ding{173} if $q_{nj} < 0$, then the nodes are connected in anti-parallel way
($Out_n\rightarrow IN^-_j$, $Out_j\rightarrow IN^-_n$). \ding{174} if
$q_{nj} = 0$, nodes are disconnected.

Note that, the original
BRIM design has a virtual ground $(\frac{V_{dd} + V_{ss}}{2})$ on the input 
terminal of nodes. Relative to this virtual ground, the polarity
of the nodal voltages represent the $\{\pm 1\}$ Ising spins. However, with
$\{0,1\}$-bit representation, we instead use a true ground.
Using this architecture as a baseline, we now add hardware support for cubic
interactions.

\subsection{Cubic Interactions}
\label{ssec:overview}

Motivated by observation 1 in Sec.~\ref{ssec:cubic_vs_qubo}, we would prefer
native support of cubic interactions. In principle, this is not difficult.
Indeed, any objective function can be supported in a straightforward manner
by a dynamical system. We simply set the system to be a gradient 
system~\cite{strogatz.book94} and let the objective function be the potential.
More concretely, let us rewrite the 3-SAT objective function (Eq.~\ref{eq:pubo_3sat})
by collecting linear, quadratic, and cubic terms, and substituting voltage of
a node $v_n$ as a continuous approximation of the discrete variable $x_n$:
\begin{equation}
    \begin{split}
        H = const - \sum_n l_nv_n - \sum_{n<j} q_{nj}v_nv_j - \sum_{n<j<k} c_{njk}v_nv_jv_k
    \end{split}
    \label{eq:cubic_3sat_hamil}
\end{equation}
To allow the dynamical system to naturally seek local minima of $H$, its
differential equation should be set to $-\nabla H$:
\begin{equation}
    \begin{split}
        \frac{dv_n}{dt} = -\alpha\frac{\partial H}{\partial v_n} 
        = \alpha(l_n + \sum_{j} q_{nj}v_j + \sum_{jk} c_{njk}v_jv_k)
    \end{split}
    \label{eq:cubic_3sat_diff}
\end{equation}
where $\alpha=\frac{1}{RC}$ is a constant. Therefore, the rate of change
of nodal voltage is determined by the total incoming current to that node.
The baseline architecture already supports the first two terms in the 
parenthesis of Eq. \ref{eq:cubic_3sat_diff} as discussed before. To support the
last term, we need 3 steps: \ding{172} use a multiplier to produce 
a voltage $v_{jk} = v_j\times v_k$, \ding{173} apply $v_{jk}$ across a resistor 
proportional  to $\frac{1}{c_{njk}}$, and \ding{174} feed the current
to node $n$. In practice, we make 3 modifications as we discuss next.

\subsubsection{Multiplier} A "true" analog multiplier is a rather expensive
circuit due to accuracy and linearity concerns. In our system, we
do not need high accuracy or linearity. Indeed, the differential
equation (Eq.~\ref{eq:cubic_3sat_diff}) itself is the result of approximation.
If we focus only on the discrete points of voltage rails (ground and $V_{dd}$),
then a simple AND gate is a satisfactory multiplier. By faithfully modeling 
a regular AND gate, we have
compared the system-level behavior. Microscopically, the trajectories of the
systems (one using ideal multipliers, the other AND gates) are different,
as expected. However, macroscopically, we do not see any difference in their
optimization capability.

\subsubsection{Cubic terms fan-in}
\label{sec:cub_fan_in}

\begin{wrapfigure}{r}{0.29\textwidth}
    \centering
    \vskip -2pt
    \includegraphics[width=0.29\textwidth]{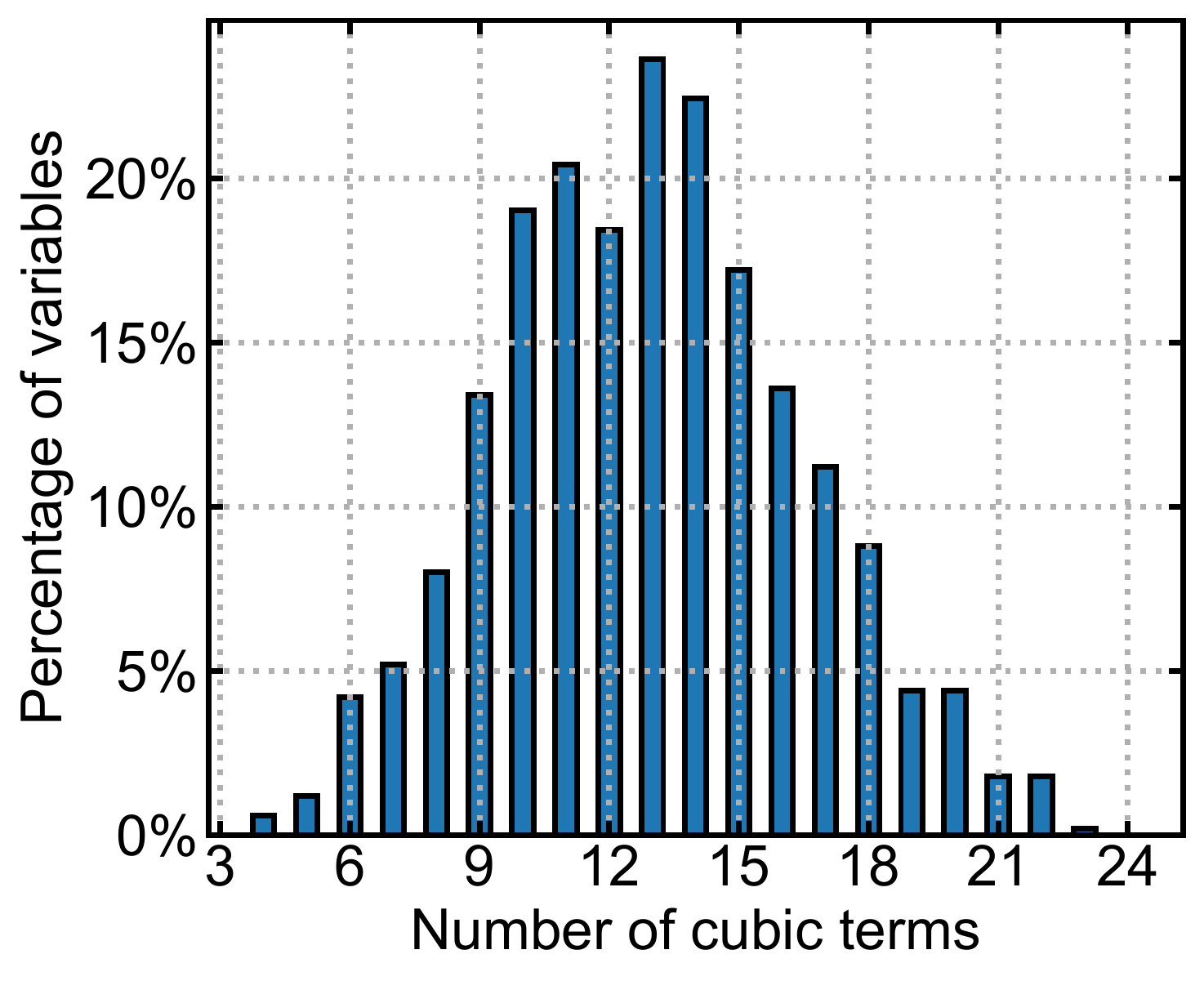}
    \vskip -5pt
    \caption{Distribution of the number of cubic terms with any variable.}
    \label{fig:dist_no_cubic}
\end{wrapfigure}
For a \emph{general} support of cubic interaction, we need $O(N^3)$ couplers. However, for 3-SAT problems, the number of cubic terms is determined by the number of clauses $M$ which, for difficult problems, is typically about $4.2N$~\cite{nishimori.book2001}. This means that \emph{on average} each variable is involved in about a dozen cubic terms -- regardless of the size of the problem. Thus, we provide a fixed degree fan-in for cubic terms. As Fig.~\ref{fig:dist_no_cubic} shows, the distribution does not have a long tail. A fan-in degree $d=20$ would work for most benchmarks we studied. Given a fan-in limit chosen by a particular machine, users can pre-process excess clauses in a number of ways such as using the KZFD rule (Eq.~\ref{eq:kzfd}).

\begin{figure}[thb]\centering
    \includegraphics[width=0.48\textwidth]{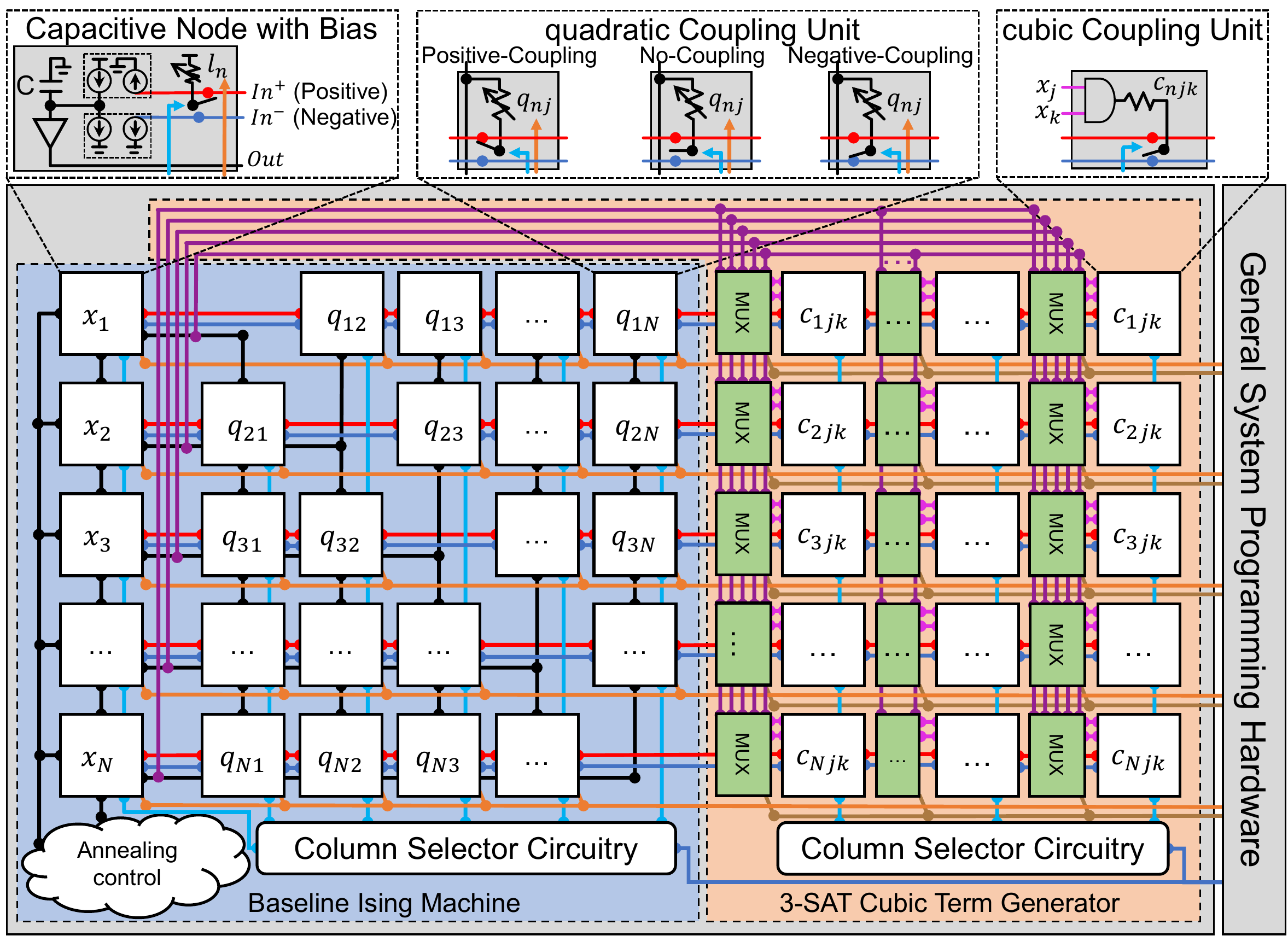}
    \caption{High-level system overview of proposed cBRIM architecture demonstrating
    how baseline BRIM system can be extended to support cubic interaction.}
    \vspace{-10pt}
    \label{fig:highlevel}
\end{figure}

The resulting architecture is shown in Fig.~\ref{fig:highlevel}.
We name this design, cubic BRIM or \emph{cBRIM} in short.
Every multiplier (AND gate) is fed by two $N$:1 multiplexers. 
This means that the cubic interactions encounter a small propagation delay
relative to the quadratic interactions. Our simulation results show
that, for our specific configuration, delays up to 3 ns are inconsequential for system performance.


\subsubsection{Coupler programmability}

Finally, another problem-specific feature of 3-SAT is that the vast majority of cubic term coefficients are $\pm 1$. For an $M$-clause 3-SAT problem, there are at most $M$ cubic terms produced by Eq.~\ref{eq:pubo_3sat}, one each by every clause with 3 literals. On rare occasions, different clauses may involve the same variables but with different negation patterns. Then, they either cancel out or add up to cubic coefficient of 2 or -- at least in theory -- even more. In such a case, that particular cubic term is simply treated as multiple instances that take up multiple couplers (fan-in). They just happen to select the same inputs for multiplication. So we eliminate the programmability on the coupler strength -- it will be fixed at 1. Instead, the programmability is on which two inputs to multiply, and the polarity of the coupling. 

\vskip 10pt

Putting it together, a 3-SAT problem in CNF will be pre-processed
via $O(3d_{max}\cdot M + N)$ time von Neumann computation (where 
$d_{max}$ is the maximum degree of the graph) to turn into a binary 
formulation (Eq.~\ref{eq:cubic_3sat_hamil})
organized as a matrix, and programmed into the hardware
column by column. For linear and quadratic terms, the programmed
values represent the coefficient of terms. For cubic terms,
the programmed value identifies the variables to be multiplied
together.


\subsection{Annealing Control}
\label{sec:annealing}

A simple yet effective annealing control is to randomly flip a spin 
with a certain probability and gradually reduce this probability~\cite{afoakwa.hpca21}. 
Motivated by observation 3 in Sec.~\ref{ssec:navigation}, we explore
alternative control that can improve the hardware efficiency in
navigating the landscape. In our 3-SAT solver architecture, there 
are at least two opportunities for improving navigational efficiency.
\ding{172} First, we show that a completely random flipping strategy can lead to 
inefficiencies which motivates us to propose a novel semantic-aware random
annealing in Sec. \ref{ssec:smart}. This requires us to
compute a metric called \emph{make count} ($\mathscr{M}_n$) discussed in Sec. 
\ref{sssec:comp_mn}
and the hardware implementation details explained in Sec. \ref{sssec:hard_impl}.
\ding{173} Second, in Sec. \ref{sssec:lock_grnd}, we propose another
optimization to stop annealing when ground state is identified.

\subsubsection{Semantic-aware randomization}
\label{ssec:smart}
\begin{figure}[htb]\centering
    \includegraphics[width=0.48\textwidth]{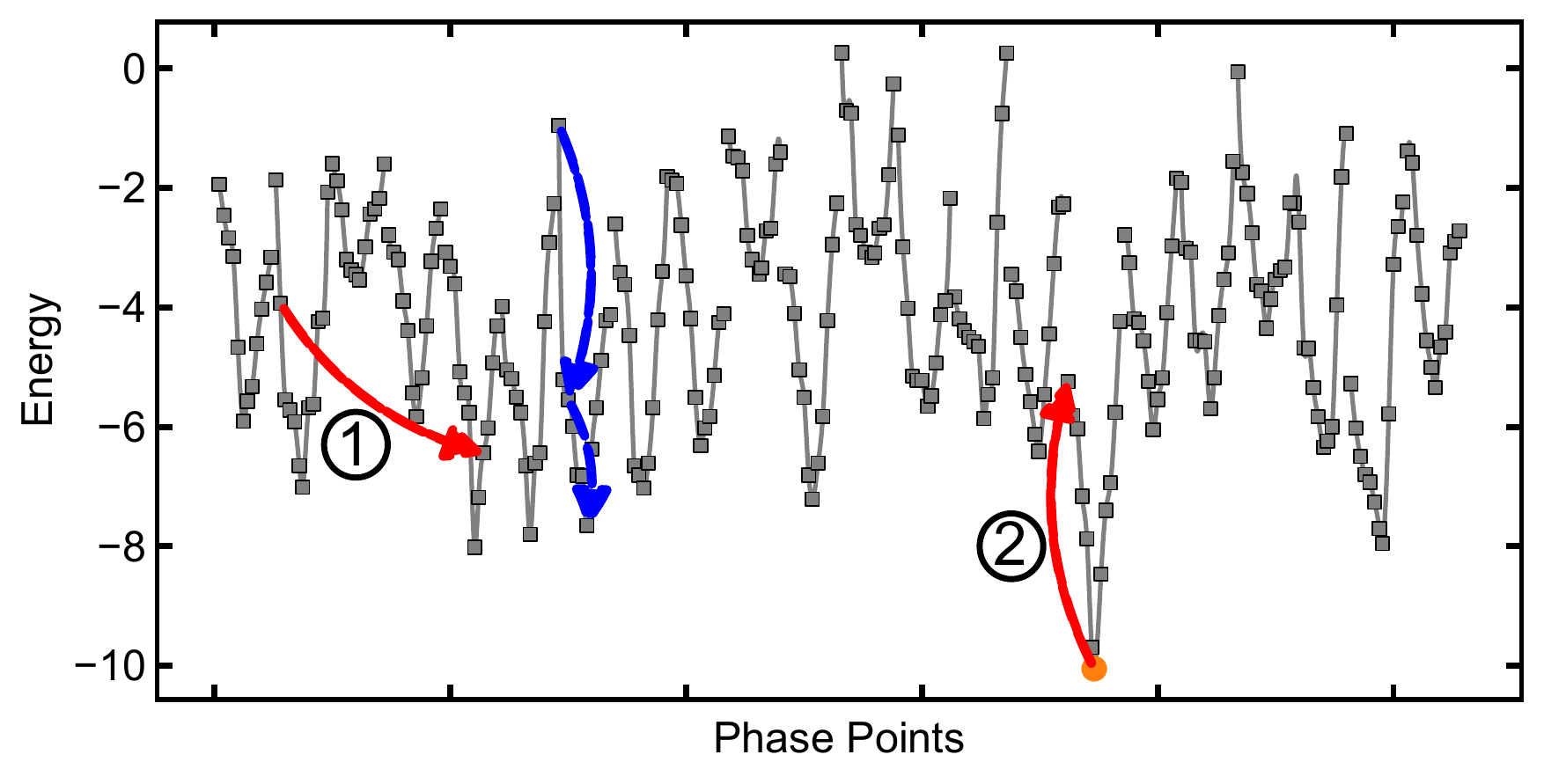}
    \vspace{-12pt}
    \caption{Illustration of landscape navigation by a dynamical system with annealing control. The horizontal axis shows all the phase point organized by their attraction region. The vertical axis shows each point's energy. State change by system dynamics is shown in blue arrows, while red arrows show random spin flips. Note that in reality the neighboring relationship between attraction regions is far more complex in a high-dimension space than can be shown in a one-dimensional plot.}
    \label{fig:landscape}
\end{figure}

To understand the first opportunity, let us see
why a baseline annealing control can be
inefficient. Without any annealing control, the dynamical system
simply evolves along trajectories determined by its differential
equations. For our type of gradient system, this inevitably leads
to (stable) fixed points~\cite{strogatz.book94}. Added random
flips allow the system to break away from the original trajectory
(Fig.~\ref{fig:landscape}). Because of its pure random nature,
such a flip can lead the system to a different attraction region
(red arrow \ding{172}) or merely to a different point inside the
same attraction region (red arrow \ding{173}). While the former
may lead to a more promising fixed point, the latter merely adds
delay.

To make this navigation more efficient, we propose a new
\emph{tanh-make-break (TMB)} heuristic which assigns flip
probability $p_n$ to each node $n$ as follows:
\begin{equation}
    \begin{split}
        p_n = \tanh (c_m\times\mathscr{M}_n)\cdot[1-\tanh (c_b\times\mathscr{B}_n)]
    \end{split}
    \label{eq:tmb}
\end{equation}
where $\mathscr{M}_n$ and $\mathscr{B}_n$ are the number of new clauses
satisfied (\emph{make count}) and unsatisfied (\emph{break count})
respectively by flipping node $n$, and $c_m > 0$ and $c_b > 0$ are
parameters. Our main motivation to use the $\tanh$ function are: 
\ding{172} easily implementable in hardware, \ding{173} directly
resembles probability without any need for normalization, and \ding{174} 
is non-linear as experimentally we found it to work better. Finding the
most optimal function is
left as future work. Thus, if a node has very high $\mathscr{M}_n$
compared to its $\mathscr{B}_n$, then it will have high chances of
getting flipped. On the contrary, a node with relatively high
$\mathscr{B}_n$ may be prohibited by the heuristic to flip.
$c_m$ and
$c_b$ determine how much weight to be assigned to $\mathscr{M}_n$
and $\mathscr{B}_n$. The main intuition is that, stochastically 
flipping nodes that satisfy more clauses makes the
landscape navigation more efficient. We shall see later that
with tuned $c_m$ and $c_b$ parameters, this indeed seems to be the
case.

We use analog circuits to implement our TMB heuristic in hardware.
In this regard, we need to compute
$\mathscr{M}_n$ and $\mathscr{B}_n$ for each node. It turns out, the
incoming current to each node is related to the difference
($\mathscr{M}_n-\mathscr{B}_n$) as shown below.
\begin{equation}
    \begin{split}
        \frac{dv_n}{dt} &= \alpha(1-2x_n)(\mathscr{M}_n-\mathscr{B}_n)
    \end{split}
    \label{eq:mb_dvdt}
\end{equation}
We omit the proof here to avoid the tedium.
We opt to compute $\mathscr{M}_n$ since
it is much easier to obtain and recover $\mathscr{B}_n$ from the incoming
current in Eq.~\ref{eq:mb_dvdt}.

\subsubsection{Computing $\mathscr{M}_n$}
\label{sssec:comp_mn}
Let us consider the Hamiltonian ($H_n$) of only those clauses involving 
node $n$. This function has some terms that contain $v_n$ (we call them
collectively $\mathscr{R}_n$) and the remaining that do not
($\mathscr{R}_{\symbol{92}n}$).
\begin{equation}
    \begin{split}
        H_n &= const. - l_nv_n - \sum_{j} q_{nj}v_nv_j - \sum_{jk} c_{njk}v_nv_jv_k\\
        &= \mathscr{R}_n + \mathscr{R}_{\symbol{92}n} = \mathscr{M}_n
    \end{split}
    \label{eq:Hn}
\end{equation}
Now, $H_n$ is just the number of unsatisfied clauses associated with 
node $n$ and hence, is just $\mathscr{M}_n$. If we notice
Eq. \ref{eq:cubic_3sat_diff}, the incoming current of each node is simply
$\frac{dv_n}{dt}=-\alpha\pdv{\mathscr{R}_n}{v_n}$. This is because only
$\mathscr{R}_n$ contains terms with $v_n$. Using this fact and applying 
Eq. \ref{eq:mb_dvdt} in Eq. \ref{eq:Hn},
\begin{equation}
    \begin{split}
    \mathscr{M}_n&=H_n = x_n\pdv{\mathscr{R}_n}{v_n} + \mathscr{R}_{\symbol{92}n} \\
    &= -\frac{x_n}{\alpha}\cdot\frac{dv_n}{dt} + \mathscr{R}_{\symbol{92}n} \\
    &= x_n(\mathscr{M}_n-\mathscr{B}_n) + \mathscr{R}_{\symbol{92}n}
    \end{split}
    \label{eq:Hn2}
\end{equation}
where $x_n$ is the quantized state of the node. While deriving Eq. 
\ref{eq:Hn2}, we use a simplification that since $x_n\in\{0,1\}$, we
can write $x_n^2=x_n$. The first term in Eq. \ref{eq:Hn2} can be 
obtained from the incoming
current and only contributes when $x_n=1$. The second
term ($\mathscr{R}_{\symbol{92}n}$) can be computed by generating a
current with voltages of participating nodes applied across extra
coupling units. As a result, we generate a current proportional to 
$\mathscr{M}_n$. With $\mathscr{M}_n$ computed, we can extract $\mathscr{B}_n$
using Eq.~\ref{eq:mb_dvdt}.

\subsubsection{Hardware implementation}
\label{sssec:hard_impl}
\begin{figure}[htb]\centering
    \includegraphics[width=0.40\textwidth]{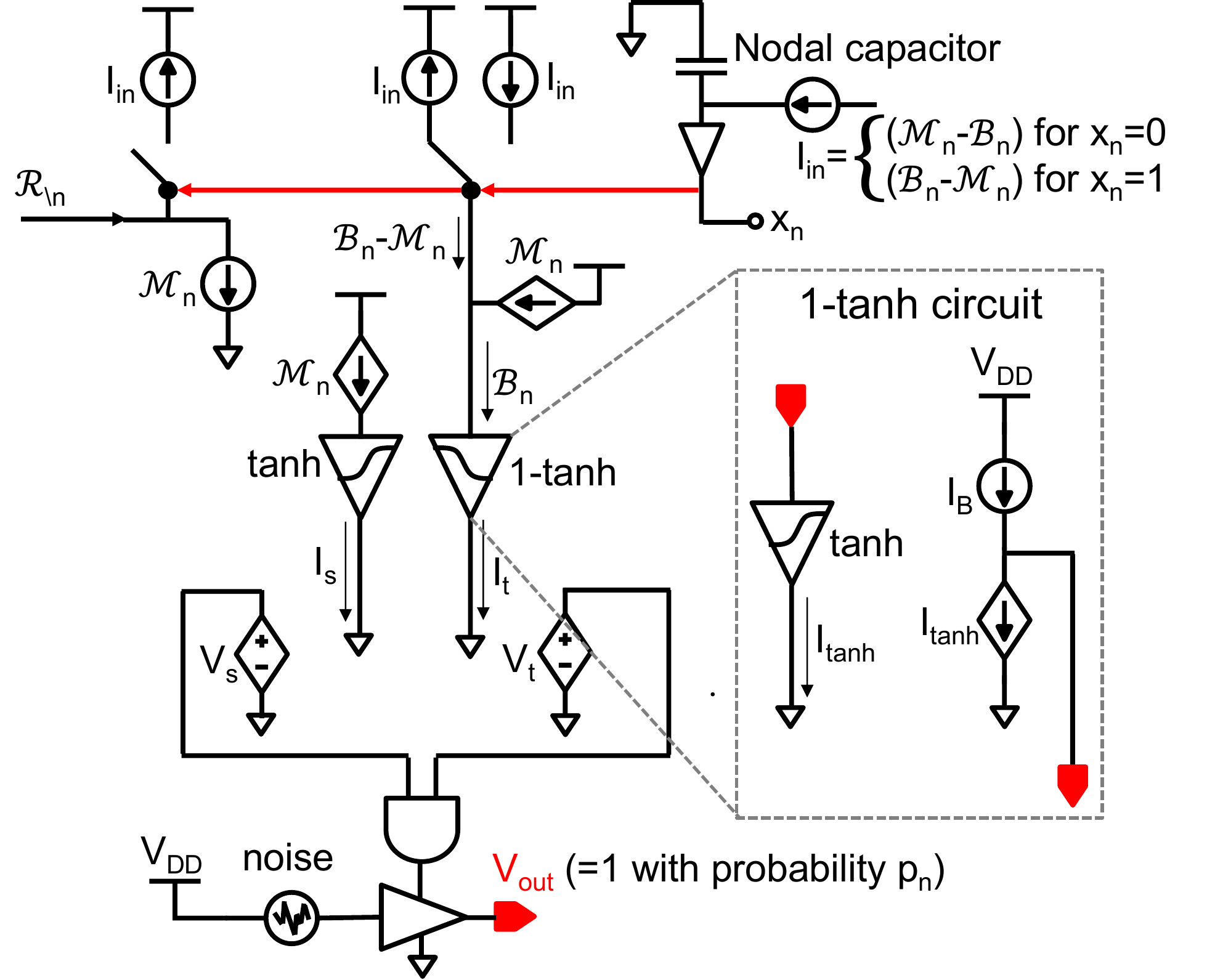}
    \caption{The CMOS circuit implementation of the heuristic.}
    \label{fig:tmb}
\end{figure}

\begin{figure}[htb]\centering
    \includegraphics[width=0.40\textwidth]{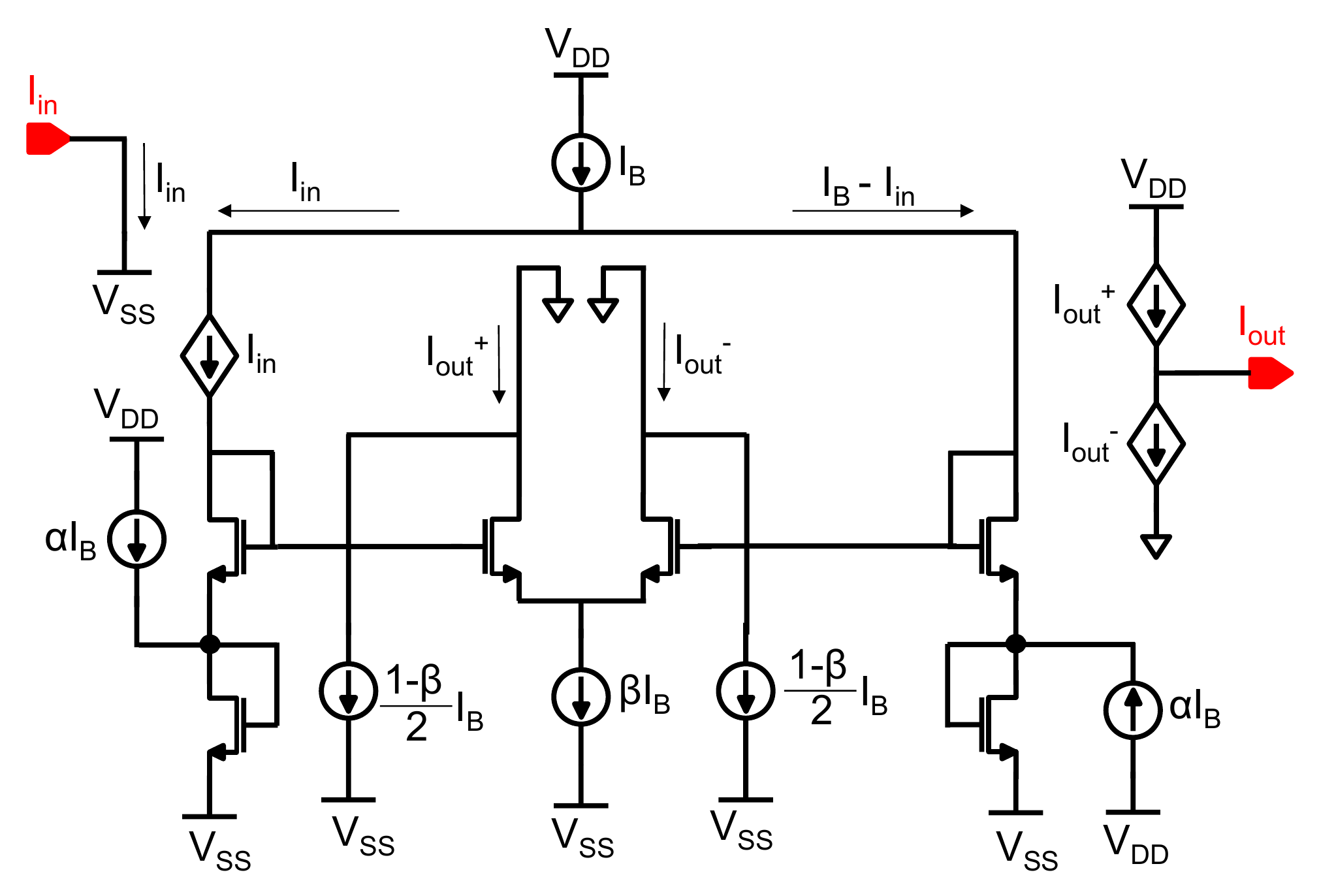}
    \caption{CMOS implementation of a single-ended $\tanh$ circuit.}
    \label{fig:tanh_ckt}
\end{figure}

With $\mathscr{M}_n$ and $\mathscr{B}_n$ represented as currents,
we can design a circuit to compute $p_n$ in Eq. \ref{eq:tmb} with approximate $\tanh$ generators and using an AND gate to approximate the multiplication. Fig. \ref{fig:tmb} shows an example implementation. The current $\mathscr{R}_{\symbol{92}n}$ is generated using coupling units not shown in the figure. Based on the quantized state of the node $x_n$, a switch is set to ensure the current flowing into the ($1-\tanh$) circuit is always $\mathscr{B}_n$ and also to get $\mathscr{M}_n$ from $\mathscr{R}_{\symbol{92}n}$. The $\tanh$ function is implemented by modifying a differential current-mode implementation shown in a recent work~\cite{cmostanhimplementation}. The resulting circuit is shown in Fig. \ref{fig:tanh_ckt}. Using this, ($1-\tanh$) is trivial to implement as shown in the zoomed-in part of Fig. \ref{fig:tmb}. The parameters $c_m$ and $c_b$ are tuned by applying a gain to the current-controlled current source that mirrors the input current in the $\tanh$ circuit. The output currents of the $\tanh$ and ($1-\tanh$) circuits, referred to as $I_s$ and $I_t$, are used to control the current-controlled voltage sources $V_s$ and $V_t$ respectively. Note that, $I_s, I_t\in[0, I_{bias}]$ because the input currents, $\mathscr{M}_n$ and $\mathscr{B}_n$ are positive. $V_s$ and $V_t$ produce voltages in the range $[0, V_{DD}]$ which are fed as inputs to an AND gate to multiply approximately. The output of AND gate controls the probability $p_n$ of a stochastic circuit that outputs $V_{out}=1$ with a probability $p_n$. This way, each node can be selected by our TMB heuristic to flip if $V_{out}=1$.

While the TMB heuristic requires extra hardware,
it is only needed per node and thus
comes with $O(N)$ cost. The system's overall
circuit complexity is dominated by that of the $O(N^2)$
couplers in practical systems.

In a fast dynamical system like BRIM, flipping a node is a stochastic mechanism to
escape local minima. It is done by "clamping" the node to a fixed polarity using a large current driver for a specific duration. 
Assuming the system started in a fixed point (local minimum), then after such a temporary 
clamp to a single node, the system will almost certainly revert back to the same fixed point. We need multiple clampings to overlap in order to escape.

By its very nature, our design flips nodes in an
asynchronous manner and allows multiple stochastically overlapping flips. 
Lengthening the duration increases the number of overlaps. 
We will study the effect of the choice of clamp duration on the solution quality in 
Sec.~\ref{sec:eval_clamp}.


\subsubsection{Latching ground state}
\label{sssec:lock_grnd}
Another inefficiency is that the dynamical system has no explicit notion 
of energy the way a software algorithm does: it cannot determine 
whether one local minimum is better than another. In other words, 
software can always remember the phase point with the best energy ever visited,
but a hardware solver can only read out the final state or
at most a finite number of snapshots of the system's trajectory.
In a pure optimization task, this is an acceptable tradeoff in 
exchange for the extreme speed and efficiency. In SAT problems,
the situation is a bit different. \ding{172} To solve the problem,
we generally need to reach the ground state (though finding the maximum
number of satisfiable clauses is useful too~\cite{koshimura2012qmaxsat}). 
\ding{173} Unlike in many other problems (\eg MaxCut), it is quite easy to  know when the ground state is reached.

To sum, the risk-reward
balance of a more explicit tracking of the energy value is shifted
quite a bit in SAT. Thus we add an additional feature: 
when all clauses are satisfied, we
simply stop adding any random flips and indicate the answer is found
(\emph{latching}).
This is particularly convenient given our TMB heuristic: the formula is satisfied
when $\mathscr{M}_n=0$ for all $N$ variables.
As we will show in the analysis, this has a non-trivial 
boost to the effective speed of the machine.

\section{Experimental Analysis}\label{sec:eval}

\subsection{Experimental methodology}
Most of the experiments are performed as a mixture of simulation and native execution.
In all the experiments, CPU software solvers like 
WalkSAT\cite{walksat} and KISSAT~\cite{kissat_21} are natively
executed on Intel Xeon Platinum 8268 CPU at 2.90 GHz.
WalkSAT is run using the \texttt{-best} heuristic also known 
as "SKC" \cite{walksat_paper} and a cutoff flips of 500,000.
KISSAT is run using the default configuration.
GPU-based solvers (ParaFrost \cite{parafrost} and AnalogSAT
\cite{gpu.analogsat}) are run on an AMD 
Ryzen 5 2600 CPU at 3.4 GHz with 16 GB of 3200 MHz DDR4 RAM and 
an NVIDIA RTX 2070 Super GPU running at stock frequencies.
The evolution of various dynamical systems is modeled by solving
differential equations using 4th-order Runge-Kutta method.
For a rough sense of the chip-based implementation, we perform a 
preliminary design and layout of components in Cadence using the
generic 45nm PDK.

For SAT benchmarks, we use those publicly
available from SATLIB~\cite{satlib} and SAT competitions. For direct comparison
with other hardware solvers, we use the same publicly available problems.
Moreover, we generate "hard" uniform random instances of varying sizes with 
$\alpha_c = 4.25$. Majority of the hardware SAT solvers only test on uniform 
random problems~\cite{amoebasat_fpga,amoebasat_fpga_original,high_order_ising_oim} 
or tiny toy problems~\cite{bashar2023oscillator}, however, it has been established
that industrial problems exhibit \emph{scale-free}
behaviour~\cite{power_law_generator, power_law_industrial}. Therefore, we 
generated some difficult industrial-like SAT
instances with 500 variables using scale-free distribution with $\alpha_c=3.29$
(1645 clauses) and power-law exponent of 2.8~\cite{power_law_generator}.


\subsection{High-level analysis}
\begin{figure}[htb]\centering
    \includegraphics[width=0.47\textwidth]{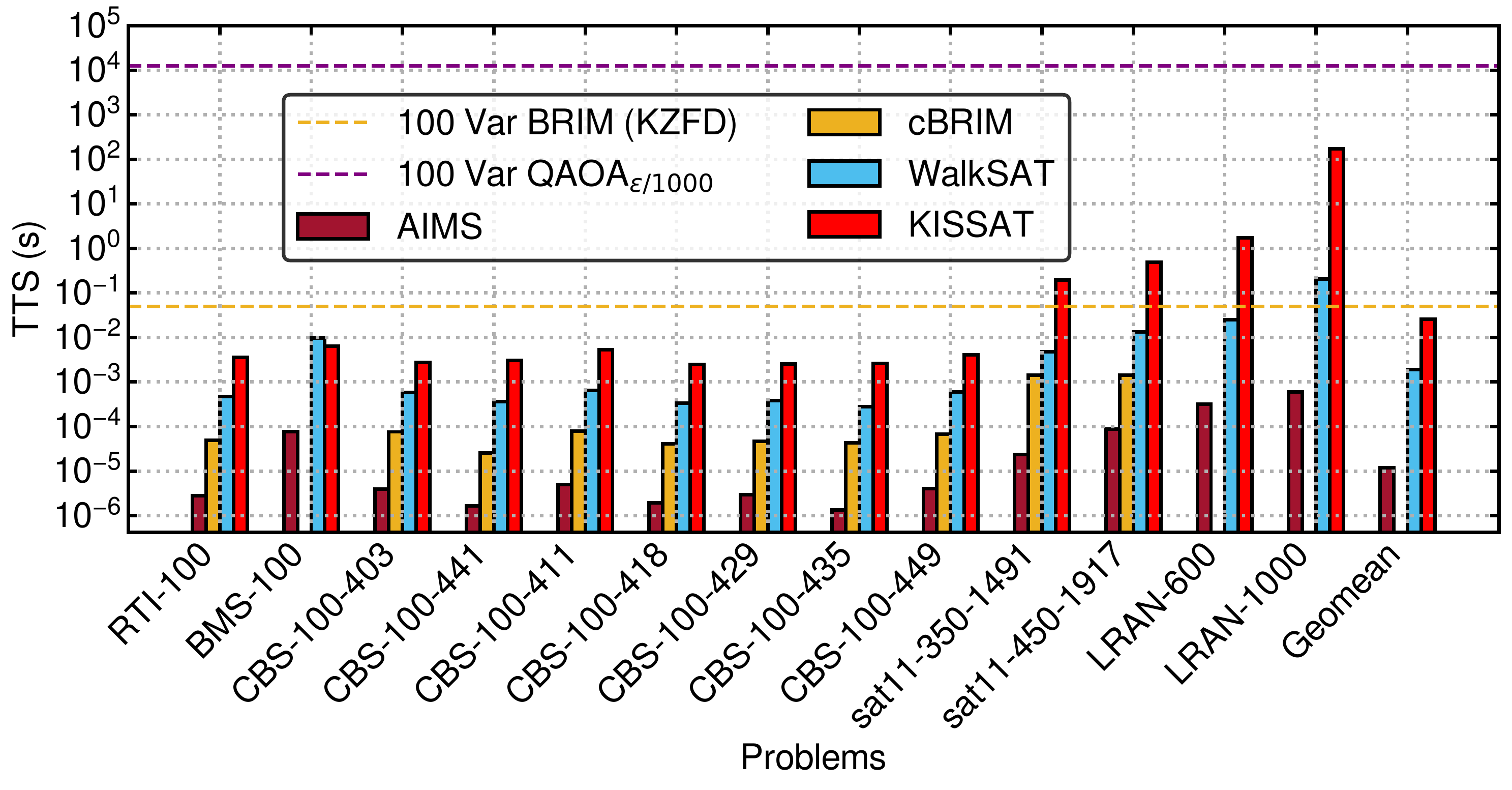}
    \vspace{-9pt}
    \caption{Time-To-Solution (s) of AIMS, cBRIM, WalkSAT and KISSAT for various benchmark suites from SATLIB and SAT 2011 competition. \emph{sat11} benchmark suites contain 4 problems each, both the \emph{LRAN} benchmarks contain a single problem and the remaining benchmarks consist of 10 problems each. Each benchmark suite is named in this format: $\langle$Benchmark$\rangle$--$\langle$Vars$\rangle$--$\langle$Clauses (if stated)$\rangle$.}
    \vspace{-9pt}
    \label{fig:tts_bench}
\end{figure}

We start with a high-level analysis comparing the architecture
with all supports discussed in Sec.~\ref{sec:arch}. We tuned the parameters
$c_m$ and $c_b$ in our TMB heuristic that works the best, in general, for all
problems. This augmented Ising machine for SAT (or AIMS for brevity) is 
first compared against state-of-the-art software solvers and cBRIM
as defined in Sec.~\ref{sec:cub_fan_in}. Note that, existing high order
Ising machines have been either evaluated with tiny 
problems~\cite{bashar2023oscillator} or with poor solution quality and no direct
report of the solver time~\cite{high_order_ising_oim}. In contrast, cBRIM is able
to find solutions with higher success rates and hence, is compared here.
Fig.~\ref{fig:tts_bench} shows the Time-To-Solution (TTS) for all the 
solvers. The plot shows the geometric mean for each benchmark suite and
also the overall TTS. 

As we can see, for these benchmarks, AIMS is consistently faster with a geometric mean speedup about 2 and 3 orders of magnitude over software solvers WalkSAT~\cite{walksat} and KISSAT~\cite{kissat_21} respectively. With contributions from our TMB heuristic, AIMS outperforms cBRIM by at least an order of magnitude. Some of the data for cBRIM are missing as it could not solve the problems in our tested annealing times. 

With the power consumption of AIMS estimated to be on the orders of 100 $mW$, energy consumption will likely be another order of magnitude better than von Neumann computing. The figure also shows the TTS of baseline BRIM using KZFD quadratization and an optimistic estimate of QAOA$_{\epsilon/1000}$ on a 100-variable problem. 
Observe that AIMS is able to solve a 1000-variable problem while still
being 2 orders of magnitude faster than baseline BRIM solving a
100-variable problem. Thus, we can see the importance of architectural
design in AIMS to reach high problem solving capabilities. 
For the software solvers, except for the BMS instances, WalkSAT is able to
outperform KISSAT. We do note that for much larger
problems, the situation might be different as is evident from the SAT
competitions \cite{kissat_21}. The case for QAOA is not good as it is
several orders of magnitude slower than even the software solvers.



Table \ref{tab:tts_hard} shows a comparison of AIMS with two other state-of-the-art hardware SAT
solvers: HW AmoebaSAT~\cite{amoebasat_fpga_original} and BRWSAT~\cite{brw_mtj}.
The benchmarks shown here are the only ones reported by the references and hence, can be
directly compared. Based only on the overlapping benchmarks, AIMS has
a geometric speedup of 7.4x and 10.3x over HW AmoebaSAT and BRWSAT, respectively.
Note that HW AmoebaSAT scales rather poorly as TTS increases 3 orders from
problem sizes of 50 variables to 225 variables.

\begin{table}[htb]\centering\footnotesize
\caption{Comparison of TTS for various hardware solvers in units of $\mu s$.
\textbf{\#V}, \textbf{\#C}: the number of variables and clauses, respectively.}
\label{tab:tts_hard}
\scalebox{0.9}{
\hspace{-6pt}
\setlength{\tabcolsep}{4.5pt}
 \begin{tabular}{| l| r | r | r | r | r | r |}
\hline
\textbf{Benchmark}  & \textbf{\#V} & \textbf{\#C} & \textbf{HW AmoebaSAT}  & \textbf{BRWSAT} & \textbf{AIMS}  \\ \hline\hline
uf50-0100       & $50$  & $218$     & $4.126$   & $-$       & $5.829$ \\ \hline
uf50-0410       & $50$  & $218$     & $4.360$   & $-$       & $1.329$ \\ \hline
uf50-0767       & $50$  & $218$     & $8.300$   & $-$       & $4.059$ \\ \hline
uf100-0285      & $100$ & $430$     & $356$     & $-$       & $73.81$ \\ \hline
uf150-0100      & $150$ & $645$     & $1832$    & $-$       & $73.16$ \\ \hline
uf225-028       & $225$ & $960$     & $3078$    & $-$       & $10.41$ \\ \hline
sat11-350-p11   & $350$ & $1491$    & $-$       & $200$     & $39.54$ \\ \hline
sat11-350-p23   & $350$ & $1491$    & $-$       & $1289$    & $44.56$ \\ \hline
sat11-350-p75   & $350$ & $1491$    & $-$       & $280$     & $22.81$ \\ \hline
sat11-350-p98   & $350$ & $1491$    & $-$       & $50$      & $7.389$ \\ \hline
sat11-450-p1    & $450$ & $1917$    & $-$       & $80$      & $6.448$ \\ \hline
sat11-450-p11   & $450$ & $1917$    & $-$       & $1912$    & $577.1$ \\ \hline
sat11-450-p14   & $450$ & $1917$    & $-$       & $4650$    & $786.2$ \\ \hline
sat11-450-p78   & $450$ & $1917$    & $-$       & $353.1$   & $8.082$ \\ \hline
\end{tabular}
}
\end{table}

\begin{figure}[htb]
    \centering
    \includegraphics[width=0.36\textwidth]{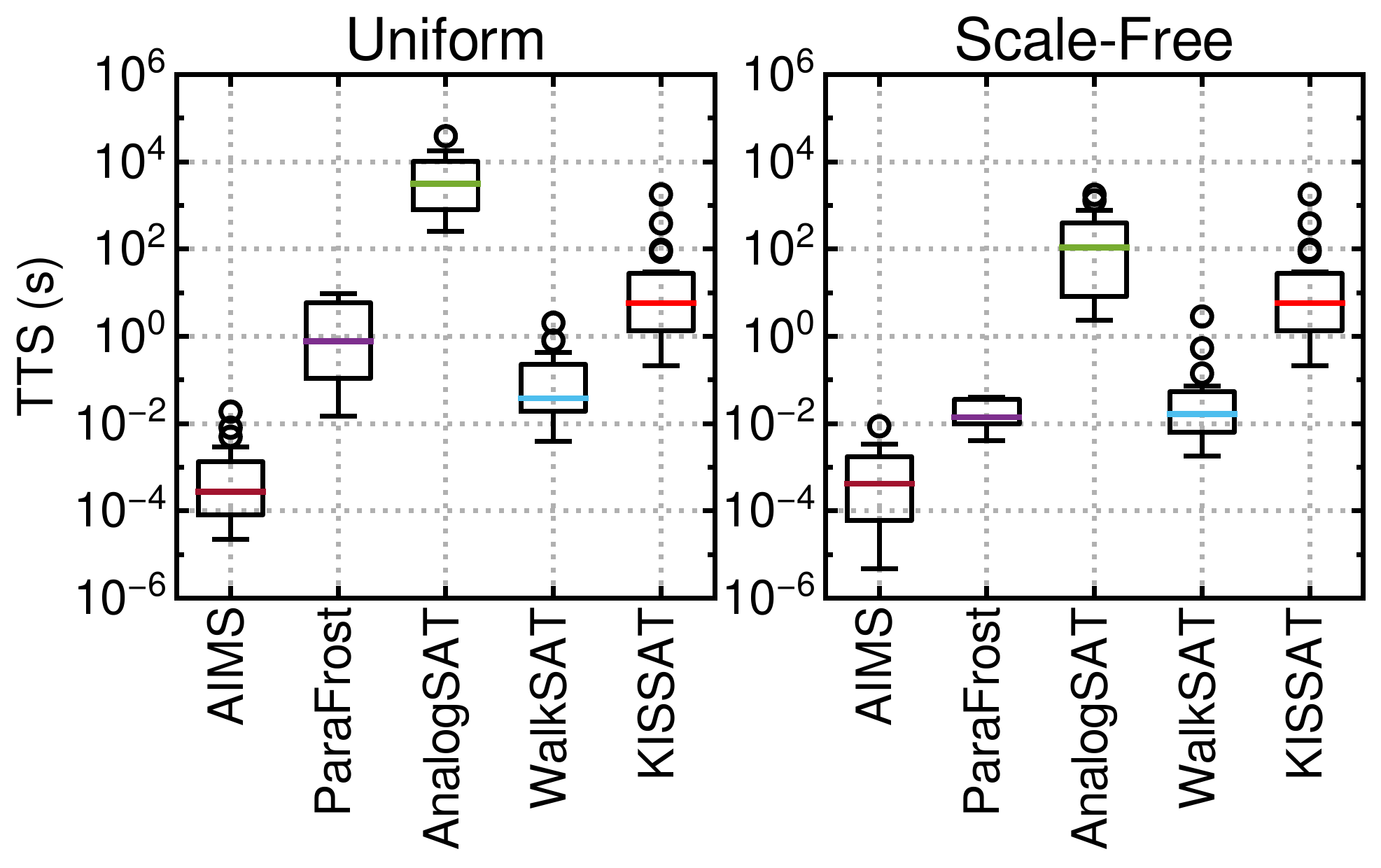}
    \caption{Box plot showing distribution of TTS for AIMS, GPU-based solvers: ParaFrost and AnalogSAT, and software SAT solvers: WalkSAT and KISSAT. All solvers run 20 problems each with [Left] uniform random and [Right] industrial-like scale-free distribution of 500 variables.}
    \label{fig:gpu}
\end{figure}

We also compare against GPU-based solvers.
Fig. \ref{fig:gpu} shows the box plot distribution of TTS for 
AIMS compared with two state-of-the-art GPU-based SAT solvers:
ParaFrost \cite{parafrost} and AnalogSAT \cite{gpu.analogsat}. The results of WalkSAT and KISSAT are 
also shown for reference. Each box expands from first quartile 
($Q_1$)
to third quartile ($Q_3$) of the results with a colored
line at the median. The whiskers extend 1.5x of the inter-quartile
distance ($Q_3-Q_1$) and the circle markers show outliers past
the end of the whiskers. 

We can observe that AIMS outperforms the better performing GPU-based solver, 
ParaFrost, by about 3 orders of magnitude 
for uniform random problem (left) and by 62x for industrial-like scale-free problems (right).


\subsection{In-depth analysis}
\label{sec:indepth}

\subsubsection{Navigational efficiency}

\begin{figure}[htb]
    \centering
    \vspace{-10pt}
    \includegraphics[width=0.42\textwidth]{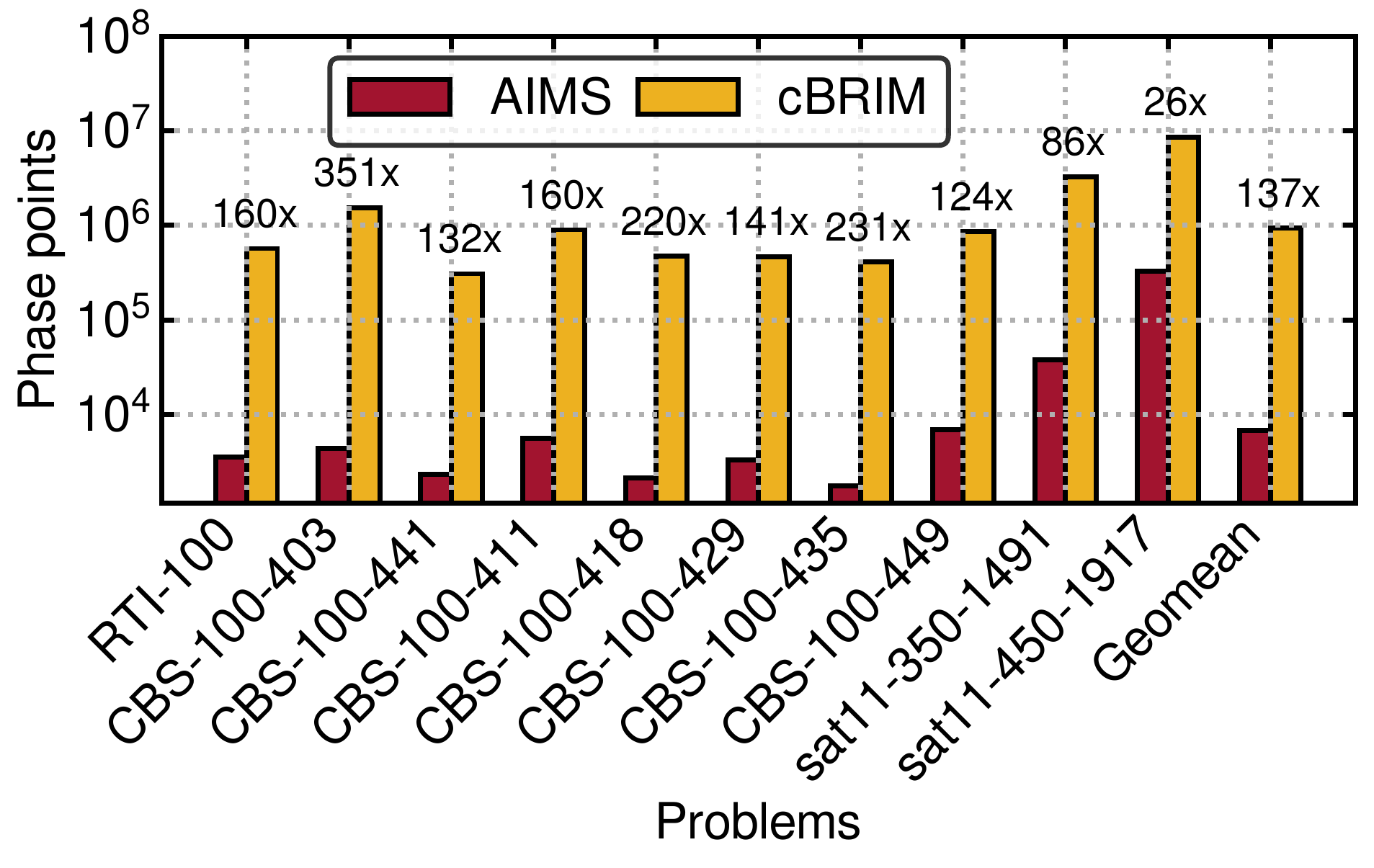}
    \vspace{-10pt}
    \caption{Phase points visited by AIMS and cBRIM for various benchmark suites
    from SATLIB and SAT competition.}
    \label{fig:phase}
\end{figure}

First, we look at the navigational efficiency of AIMS. Fig.~\ref{fig:phase} shows the average number of phase points visited to reach SAT solution by AIMS and by cBRIM. Note here that the phase points visited is also adjusted by solution probability as in TTS (Eq.~\ref{eq:tts}). 
AIMS is consistently more efficient than cBRIM as it could achieve a satisfying solution by visiting about 137x less phase points. Note again, in some
benchmarks, cBRIM fails to find a solution, suggesting that it may require
search through many more phase points. These benchmarks are not included
here.

\subsubsection{Impact of ground state latching}

\begin{figure}[htb]
    \centering
    \includegraphics[width=0.46\textwidth]{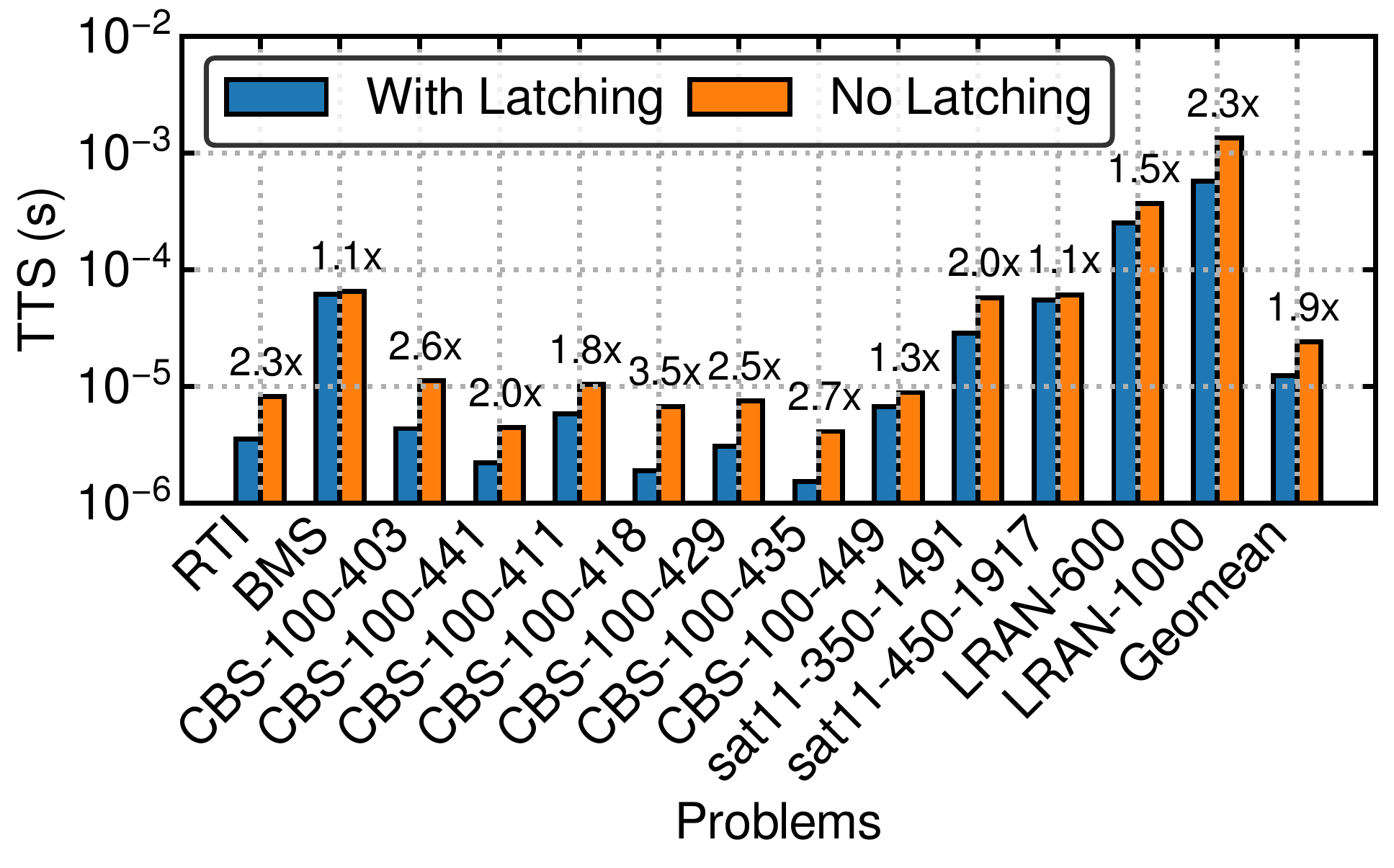}
    \caption{Results of AIMS with and without ground state latching.}
    \label{fig:latch}
\end{figure}

Next, we compare TTS with and without the result-latching logic discussed in Sec.~\ref{sssec:lock_grnd}. Fig.~\ref{fig:latch} shows the best TTS AIMS can obtain with and without latching. We observe that with latching, the results are generally better as the system terminates immediately when a solution is found, rather than to run until the user defined cutoff time. On average, we find that with this logic, TTS improves by 1.9x compared to the best TTS obtained without latching. We note that beyond this speed improvement, the latch logic is likely to be of important \emph{practical} value. It reduces the reliance on the user to set a judicious annealing time.

\donotshow{
 \begin{wrapfigure}{r}{0.25\textwidth}
    \vskip -10pt
    \centering
        \includegraphics[width=0.25\textwidth]{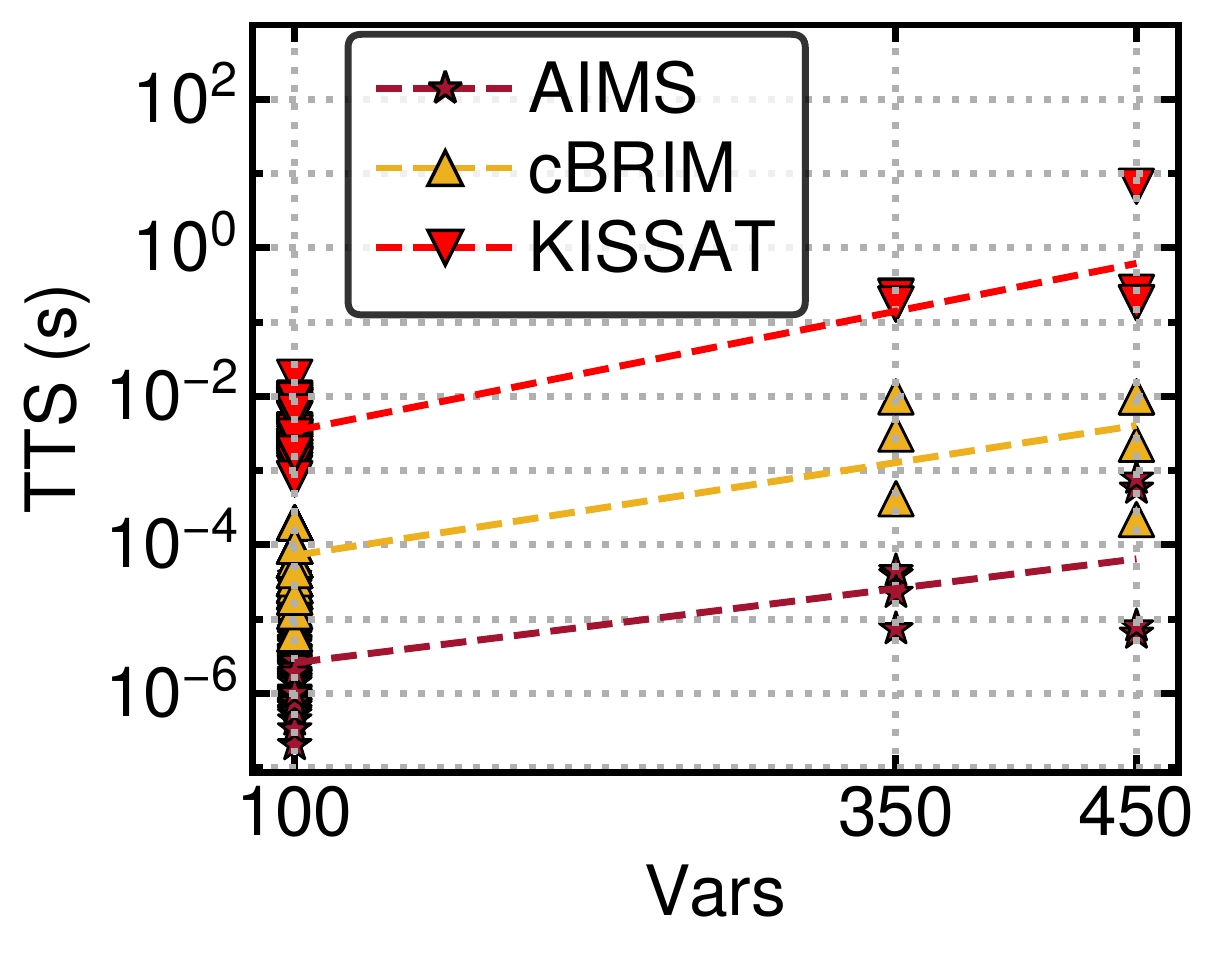}
    \vskip -10pt
        \caption{The trend of TTS for various solvers.}
        \label{fig:trend}
 \end{wrapfigure}

The navigational efficiency discussed above together with inherent speed advantage of AIMS leads to good TTS as shown in Fig~\ref{fig:trend}. The dashed lines show the linear fit of the individual data points in log scale.
Overall, the combined effect of the architectural support in AIMS (cubic interaction and navigational efficiency through heuristics and result-latching) is 3-4 orders of magnitude speedup over the baseline BRIM and 1 order speedup over cBRIM. The result underlies the importance of architectural design even for physics-based computing substrates.
}
 
\subsubsection{Relationship between clamp time and performance}
\label{sec:eval_clamp}
\begin{figure}[htb]\centering
    \includegraphics[width=0.45\textwidth]{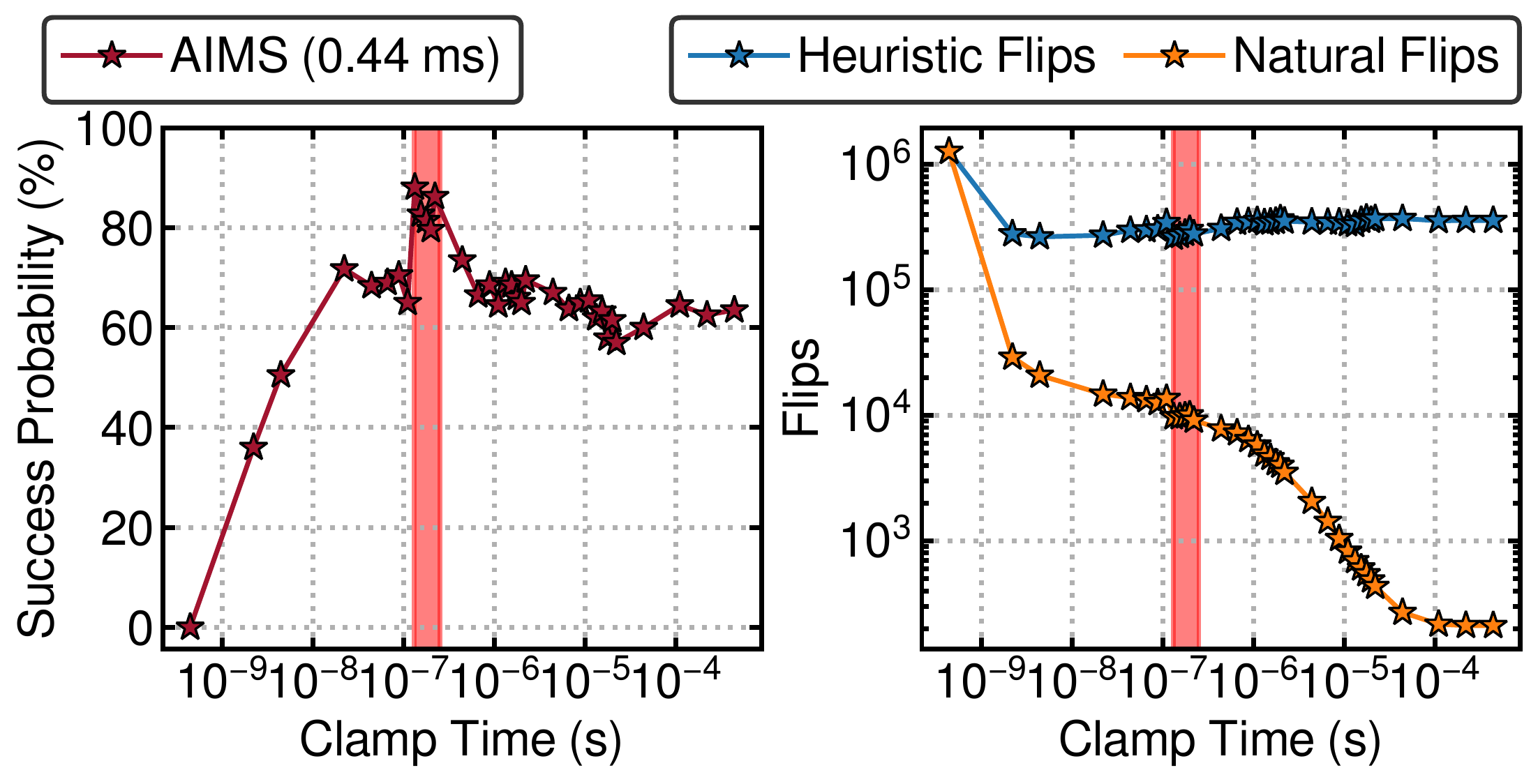}
    \caption{[Left] Success probability (\%) of AIMS while solving
    20 instances of 500-variable problems with a cutoff time of 0.44 ms
    and increasing node clamp time. [Right]
    The number of heuristic and natural flips as node clamp time 
    increases. The vertical red region shows the optimal clamp times.}
    \label{fig:clamp}
\end{figure}

In Sec. \ref{sssec:hard_impl}, we discussed about asynchronous flips with
node "clamping" which is the time until which a node's voltage
is held constant after a flip. Fig. \ref{fig:clamp} [Left] 
shows the success probability of AIMS while solving 20 instances
of 500-variable problems with increasing node clamp time and constant
annealing time of 0.44 ms. From the figure, we can observe that
too short clamp times result in poor solution quality, most likely
because the system did not get enough time to react to the flips.
There are some optimal clamp times (shown by the "red" vertical region)
resulting in the best
success probability. Fig. \ref{fig:clamp} [Right] shows the 
number of flips done by our TMB heuristic and the "natural" flips
done solely by the dynamical system. We notice that the 
amount of natural flips is very high for very short clamp time
that results in poor solution quality. This number reduces as clamp time
increases with the optimal region having roughly $10^4$ natural
flips. The heuristic flips increases slightly after a rapid
drop initially and are much higher than natural flips.


\subsubsection{Device variation and noise}\label{ssec:non_ideality}

We now analyze the effect of device variation on the performance of AIMS.
The variations are
obtained from Cadence by running a random Monte Carlo sampling in the Virtuoso Analog Design Environment XL. This sampling applies device mismatch and process variation, and is performed on a DC analysis by sweeping the input and measuring the output to obtain a transfer function. 1000 samples are taken for each circuit block, yielding 1000 curves relating the output to the input. These are then randomly sampled on a device-by-device basis in the behavioral model to estimate the impact on performance.

\begin{figure}[htb]
    \centering
        \includegraphics[width=0.36\textwidth]{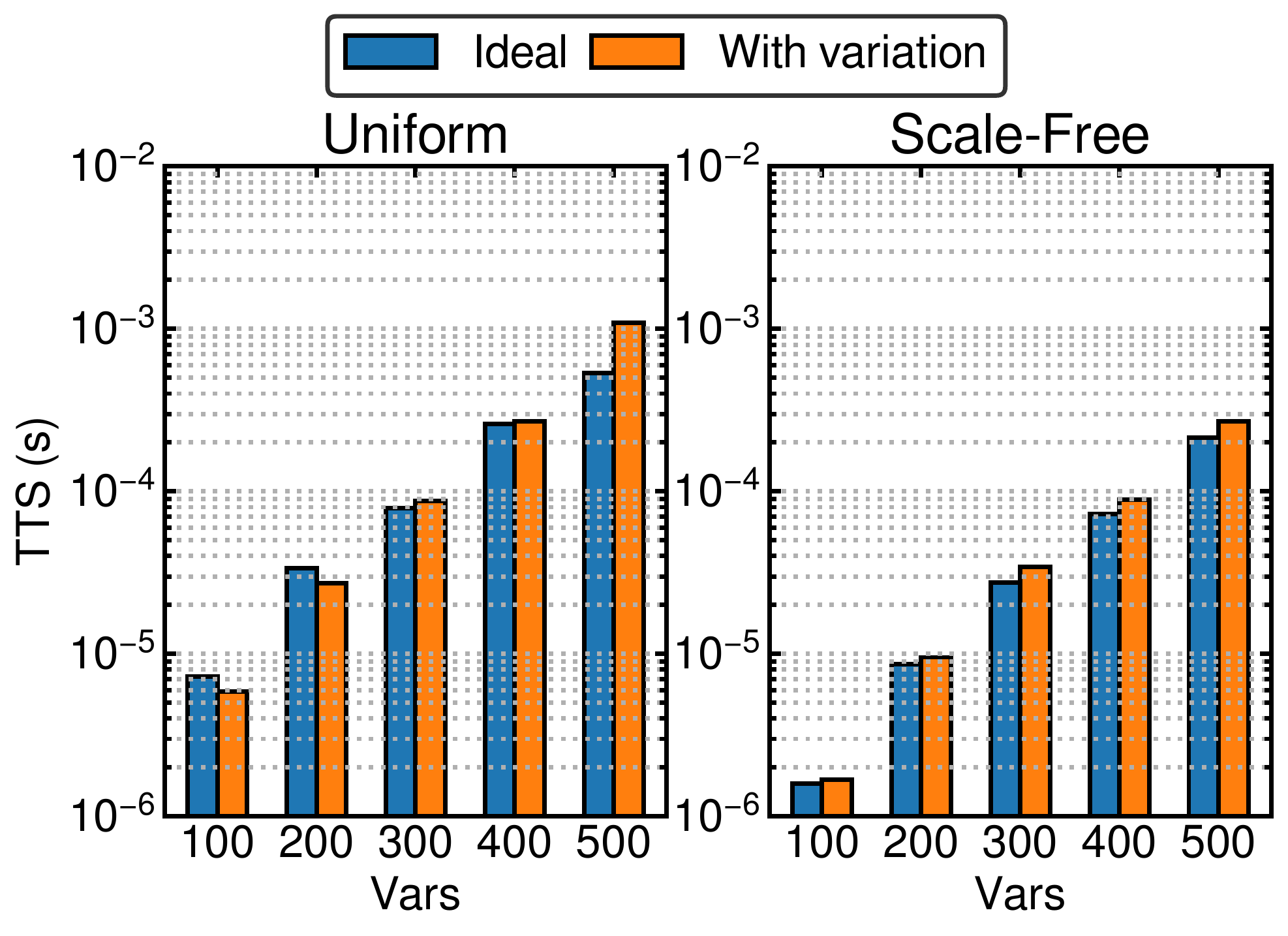}
        \caption{TTS of 3-SAT problems with different sizes
        for AIMS with ideal devices and devices with variation. Each size
        consist of 50 problems.}
        \label{fig:dev_var}
\end{figure}

Fig.~\ref{fig:dev_var} shows the TTS of AIMS with ideal devices and 
devices with variation as problem size increases for both [left] uniform 
random and [right] scale-free distribution with power-law exponent 2.8. The cutoff times are set the 
same for both the ideal and device variation runs. We can make two key
observations from the plots. First, there is 
a graceful degradation of performance and impact of
device variation is still minimal for AIMS with a capacity of 500 nodes. Second, in the case of uniform random distribution, smaller problems
(100 to 200 variables) performed better with device variation which 
implies that our TMB heuristic may be improved with further tuning.

Thermal noise analysis has only been performed on small scale systems of 50 nodes due to excessive simulation delays. A 3\% Gaussian additive noise has no appreciable impact on performance. We note that unlike quantum computing, electronic Ising machines such as AIMS are much less susceptible to noise than quantum computers where noise fundamentally disrupts the working principle. For classical annealers, circuit noise may even be harnessed to help produce needed random perturbation in annealing.

\section{Conclusions}\label{sec:conclusion}

With diminishing speed of improvements for conventional computing systems, non-von Neumann systems such as Ising machines are receiving increased attention. While all combinatorial optimization problems 
can be targeted by Ising machine in theory, 
we have shown in this paper, using SAT as a case study, 
that there are many practical hurdles with such systems:
\ding{172} QUBO reduction of cubic terms in the SAT objective function creates significant search-space bloat; \ding{173} Problem-agnostic annealing leads to poor navigational efficiency relative to state-of-the-art software solvers. 


Fortunately, these issues can be overcome with additional architectural support, at least for electronic Ising machines such as BRIM~\cite{afoakwa.hpca21}. We have presented the design details to support cubic interaction natively in BRIM. More importantly, we have proposed a novel, semantic-aware annealing control along with a ground state latching mechanism. With the additional architectural support, the resulting augmented Ising machine for SAT (AIMS) can achieve orders of magnitude speedup over state-of-the-art SAT solvers. 


Overall, our study suggests that Ising machines offer a powerful
substrate to explore novel non-von Neumann computing. At the same time,
additional architectural support may be crucial to truly unlock 
their performance potential. 

\bibliographystyle{IEEEtranS}
\bibliography{main}
\end{document}